\documentclass[11pt]{article}

\usepackage[final]{acl}

\usepackage{times}
\usepackage{latexsym}

\usepackage[T1]{fontenc}

\usepackage[utf8]{inputenc}

\usepackage{microtype}

\usepackage{inconsolata}

\usepackage{graphicx}
\usepackage{algorithm}
\usepackage{algpseudocode}
\usepackage{textcomp}
\usepackage{newfloat}
\usepackage{listings}
\usepackage{booktabs}
\usepackage{amsmath}
\usepackage{amsfonts}
\usepackage{xcolor}
\usepackage[table]{xcolor}
\usepackage{multirow}
\usepackage{makecell}
\usepackage{subcaption}
\usepackage{pifont}
\usepackage{tikz}

\usepackage{soul} 

\definecolor{BenchGreen}{RGB}{3,151,92}
\definecolor{BenchRed}{RGB}{201,52,52}
\definecolor{BenchOrange}{RGB}{255,129,3}

\newcommand{\cmark}{\textcolor{BenchGreen}{\ding{51}}}
\newcommand{\xmark}{\textcolor{BenchRed}{\ding{55}}}
\newcommand{\pmark}{\textcolor{BenchOrange}{\raisebox{0.1ex}{$\ominus$}}}
\newcommand{\gain}[1]{\textcolor{BenchGreen}{\raisebox{-0.45ex}{\scalebox{0.72}{$\uparrow$#1}}}}
\newcommand{\drop}[1]{\textcolor{BenchRed}{\raisebox{-0.45ex}{\scalebox{0.72}{$\downarrow$#1}}}}

\title{RuleWeaver: Benchmarking Rule-Centered Scenario Reasoning for Large Language Models}

\author{Bohan Yu$^{1,2}$\thanks{\,\,\,Equal contribution.}, Shi-Yang Li$^{1,2}$\footnotemark[1], Pengfei Cao$^{2,3}$\thanks{\,\,\,Corresponding authors.}, Jun Zhao$^{2,3}$, Kang Liu$^{2,3}$\footnotemark[2] \\
         $^1$School of Advanced Interdisciplinary Sciences, University of Chinese Academy of Sciences \\
  $^2$The Key Laboratory of Cognition and Decision Intelligence for Complex Systems,\\ Institute of Automation, Chinese Academy of Sciences \\
  $^3$School of Artificial Intelligence, University of Chinese Academy of Sciences \\
  \texttt{\{yubohan2025,lishiyang2026\}@ia.ac.cn \quad \{pengfei.cao,jzhao,kliu\}@nlpr.ia.ac.cn}}

\begin{document}
\maketitle

\begin{abstract}
Large language models (LLMs) are increasingly applied to specialized domains, where effective use of domain expertise often requires reasoning over complex rules in concrete scenarios. However, existing benchmarks only partially evaluate this capability, as they either focus on output-level instruction constraints or overlook the distinct roles that rules play in scenario reasoning. To address these gaps, this paper introduces \textbf{RuleWeaver}, a benchmark construction framework for evaluating rule-centered scenario reasoning. 
RuleWeaver starts from corpus-derived IF-THEN \textit{Meta Rules}, progressively augments them into complex rules, and composes these rules into rule-centered scenario QA instances.
Beyond final-answer correctness, RuleWeaver further supports process-level evaluation through rubric-based answer quality, rule recall, and rule precision. Experiments on 11 representative LLMs show that current models still struggle with complex rule-centered scenario reasoning, with even the best-performing model achieving only around 50\% of the maximum rubric score. We make our code and dataset available here: \href{https://github.com/SharkSpicy-NLP/RuleWeaver}{RuleWeaver}.
\end{abstract}

\section{Introduction}
As large language models (LLMs) demonstrate strong capabilities in text understanding, reasoning, and generation~\cite{openai2024gpt4technicalreport,zhao2023survey}, applying them in specialized domains has become increasingly important and urgent.
In such domains, it is essential to effectively leverage domain expertise, especially when this expertise is represented as explicit rules~\cite{guha2023legalbenchcollaborativelybuiltbenchmark,shen2024taskbenchbenchmarkinglargelanguage}. 
For example, under a refund policy, a model may need to decide whether a customer can receive a refund by checking rules such as ``\textit{items can be refunded within 7 days}'', ``\textit{opened items cannot be refunded}'', and ``\textit{defective items can still be refunded}''.
This example illustrates that rules can play different roles in concrete scenarios, including setting conditions, imposing prohibitions, and specifying exceptions.
In such contexts, effective model behavior requires more than producing plausible responses: models must identify rules relevant to a concrete scenario, apply them under stated constraints, and justify decisions through explicit rule-based reasoning. We refer to this capability as rule-centered scenario reasoning.

\begin{figure}[t]
    \centerline{\includegraphics[scale=0.32]{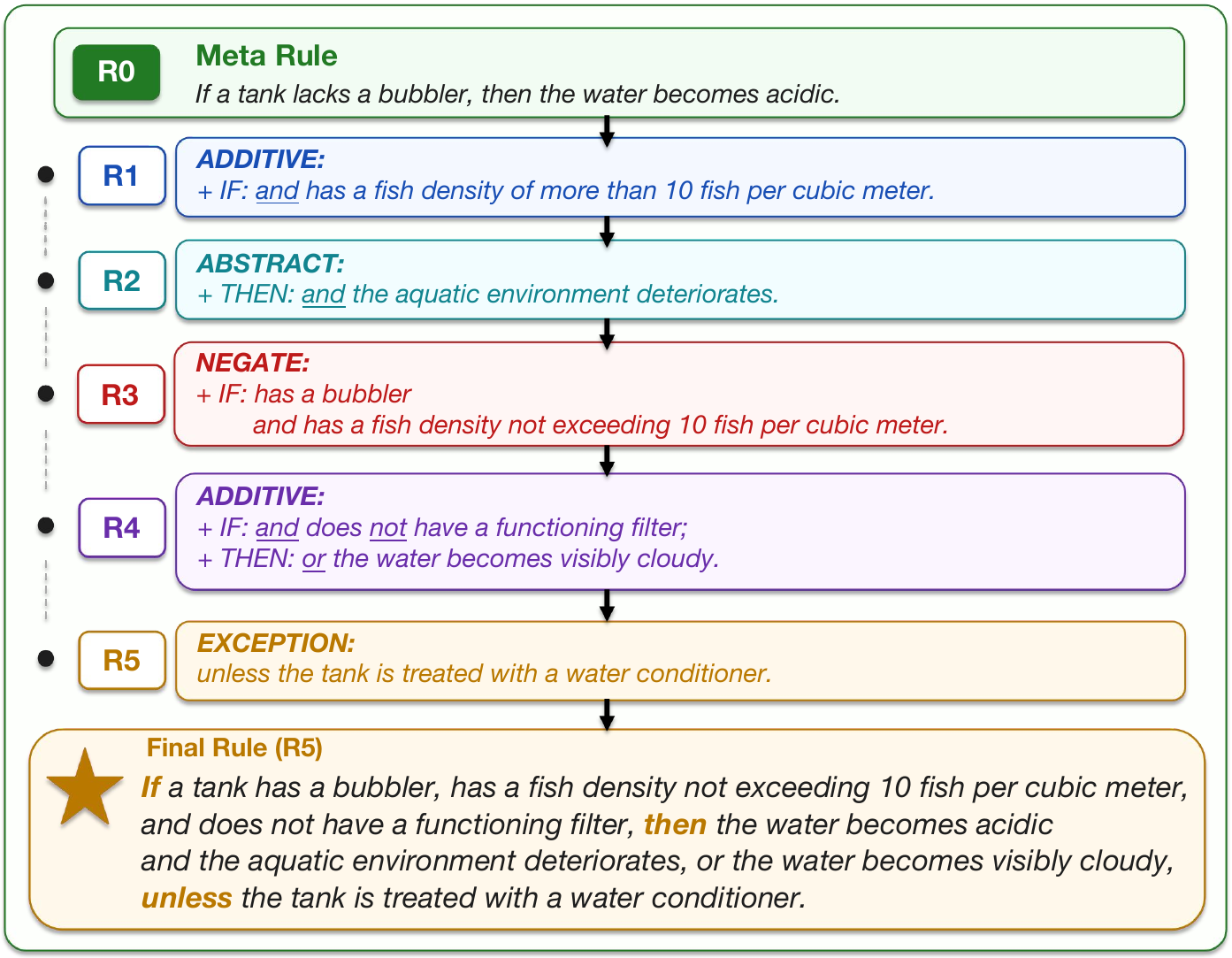}}%
    \caption{Progressive rule construction: an example of transforming a WikiHow~\cite{koupaee2018wikihowlargescaletext} Meta Rule into a complex rule through five-round semantic enhancement.}
    \label{fig:progressive_rule_construction}
\end{figure}

However, existing benchmarks, such as those for instruction following and logical reasoning, capture only part of this capability.
In specific, instruction-following datasets~\cite{jiang-etal-2024-followbench,wizardLM,qin-etal-2024-infobench,sun-etal-2025-beyond-instruction} evaluate whether models can understand constraints embedded in natural-language instructions, such as output format and content style.
Although these constraints may be expressed in different ways, they primarily prescribe what outputs should be produced or presented rather than guide reasoning or decision-making processes.
In contrast, logic-reasoning datasets~\cite{saparov2023prontoqa,han2024folionaturallanguagereasoning,parmar2024logicbenchsystematicevaluationlogical,patel2024multilogievalevaluatingmultisteplogical,zhou2025rulearenabenchmarkruleguidedreasoning,qi2025proverqa} evaluate the capability required by rule-centered scenario reasoning under the condition that the relevant rules are provided in advance. However, they mainly evaluate whether models can derive conclusions from the given rules, without sufficiently distinguishing the different roles that rules may play in realistic scenarios, such as exceptions, conflicts, or priorities.
Consequently, current benchmarks provide limited support for evaluating whether LLMs can identify relevant rules, understand their roles and dependencies, and apply them to answer questions grounded in concrete scenarios.

To address these gaps, this paper introduces \textbf{RuleWeaver}, a benchmark construction framework for evaluating LLMs on complex rule-centered scenario reasoning. The basic idea is to address two challenges: (1) building fine-grained rules with different roles, and (2) generating concrete scenarios that require those rules to be identified and applied. Specifically, for the first challenge, RuleWeaver starts from \textit{Meta Rules}, defined as atomic IF-THEN rules with a single condition and a single outcome, extracted from diverse real-world corpora spanning policy reports, procedural guides, contractual clauses, and narrative contexts. This design ensures that the Meta Rules are derived from naturally occurring rule expressions across different domains.
RuleWeaver then progressively augments each Meta Rule into a \textit{rule group} of complex variants through six semantic types: \textbf{ABSTRACT}, \textbf{ADDITIVE}, and \textbf{NEGATE} capture underspecification, supplementary constraints, and semantic reversal, while \textbf{EXCEPTION}, \textbf{CONFLICT}, and \textbf{IRONCLAD} introduce applicability changes, incompatible implications, and priority constraints.
It enables fine-grained evaluation of whether models can understand and distinguish different rules. 
Figure~\ref{fig:progressive_rule_construction} illustrates one such augmentation path: starting from a Meta Rule (R0), RuleWeaver applies sequential semantic enhancements to produce a complex rule (R5), with each step modifying the condition side, the outcome side, or both.

To address the second challenge, RuleWeaver converts the aforementioned rule groups into rule-centered scenario QA instances spanning seven predefined question types. Each instance is built around five relevant complex rules sampled from distinct rule groups under either same-source or cross-source settings.
The construction mainly includes four stages:
(1) \textit{dependency planning} selects the target question type and builds a rule dependency graph that specifies the rule applied at each reasoning step;
(2) \textit{sub-scenario generation} instantiates each planned step as a local scenario fragment with the facts needed to trigger the corresponding rule;
(3) \textit{final synthesis} merges sub-scenarios into a coherent scenario with a final question and reference outputs, including rule application annotations, a rule logic chain, and an instance-specific rubric;
and (4) \textit{iterative quality review} checks structural validity, rule coverage, information leakage, and dependency consistency.
This process ensures that the QA instances are rule-centered, with each scenario grounded in rule applications and rule-dependency chains.
Overall, RuleWeaver constructs 200 rule groups covering six semantic types and 96 scenario QA instances spanning seven question types.

For evaluation, RuleWeaver goes beyond final-answer correctness by using process annotations and instance-specific rubrics. 
Rule recall and rule precision measure whether models can identify relevant rules and apply them correctly, while rubrics assess answer quality across multiple dimensions of rule-centered scenario reasoning. 
Under this protocol, we evaluate 11 representative LLMs. The best rubric scores reach only 53.83 in the same-source setting and 50.27 in the cross-source setting, while the weakest rubric dimension achieves only a 14.2\% normalized score, revealing a clear gap between current LLM capabilities and reliable rule-centered scenario reasoning. Further analyses show that models often fail not only to identify all relevant rules, but also to apply them correctly and track dependencies among rule applications.

Our contributions are twofold:
\begin{itemize}
    \item This paper introduces RuleWeaver, a benchmark construction framework for evaluating rule-centered scenario reasoning. It builds fine-grained complex rules from corpus-derived Meta Rules and composes them into rule-centered scenario QA instances.
    \item RuleWeaver supports process-level evaluation through rubric-based answer quality, rule recall, and rule precision, revealing key weaknesses in LLMs' rule-centered scenario reasoning capabilities.
\end{itemize}

\begin{table*}[t]
\centering
\begin{minipage}[t]{0.55\textwidth}
\vspace{0pt}
\centering
\scriptsize
\setlength{\tabcolsep}{2.5pt}
\resizebox{\linewidth}{!}{
\begin{tabular}{lcccc}
\toprule
Dataset 
& \makecell[c]{Corpus-Derived\\Rules}
& \makecell[c]{Semantic\\Rule\\Typing}
& \makecell[c]{Rule-Centered\\Scenario QA with\\Process-Level Scoring}
& \makecell[c]{Multi-Source /\\Cross-Domain\\Composition} \\
\midrule
\multicolumn{5}{l}{\textbf{Instruction-Following Benchmarks}} \\
\midrule
IFEval~\cite{zhou2023instructionfollowingevaluationlargelanguage}                          & \xmark & \xmark & \xmark & \xmark \\
WizardLM~\cite{wizardLM}  & \xmark & \xmark & \xmark & \xmark \\
InfoBench~\cite{qin-etal-2024-infobench}                        & \xmark & \xmark & \xmark & \xmark \\
FollowBench~\cite{jiang-etal-2024-followbench}                        & \xmark & \xmark & \xmark & \xmark \\
ComplexBench~\cite{complex_bench}                       & \xmark & \xmark & \xmark & \xmark \\
\midrule
\multicolumn{5}{l}{\textbf{Logic-Reasoning Benchmarks}} \\
\midrule
RuleTaker~\cite{clark2020ruletaker}      & \xmark & \xmark & \pmark & \xmark \\
AR-LSAT~\cite{zhong2021arlsatinvestigatinganalyticalreasoning}        & \pmark & \pmark & \xmark & \xmark \\
ProofWriter~\cite{tafjord2021proofwritergeneratingimplicationsproofs}     & \xmark & \pmark & \pmark & \xmark \\
PrOntoQA~\cite{saparov2023prontoqa}       & \xmark & \pmark & \pmark & \xmark \\
FOLIO~\cite{han2024folionaturallanguagereasoning}          & \xmark & \pmark & \xmark & \xmark \\
LogicBench~\cite{parmar2024logicbenchsystematicevaluationlogical}     & \xmark & \pmark & \pmark & \xmark \\
Multi-LogiEval~\cite{patel2024multilogievalevaluatingmultisteplogical}   & \xmark & \pmark & \pmark & \xmark \\
RuleArena~\cite{zhou2025rulearenabenchmarkruleguidedreasoning}       & \cmark & \xmark & \cmark & \pmark \\
ProverQA~\cite{qi2025proverqa}        & \xmark & \pmark & \pmark & \xmark \\
\midrule
\textbf{RuleWeaver (Ours)} & \cmark & \cmark & \cmark & \cmark \\
\bottomrule
\end{tabular}
}
\end{minipage}
\begin{minipage}[t]{0.39\textwidth}
\vspace{-15pt}
\centering
\includegraphics[width=\linewidth]{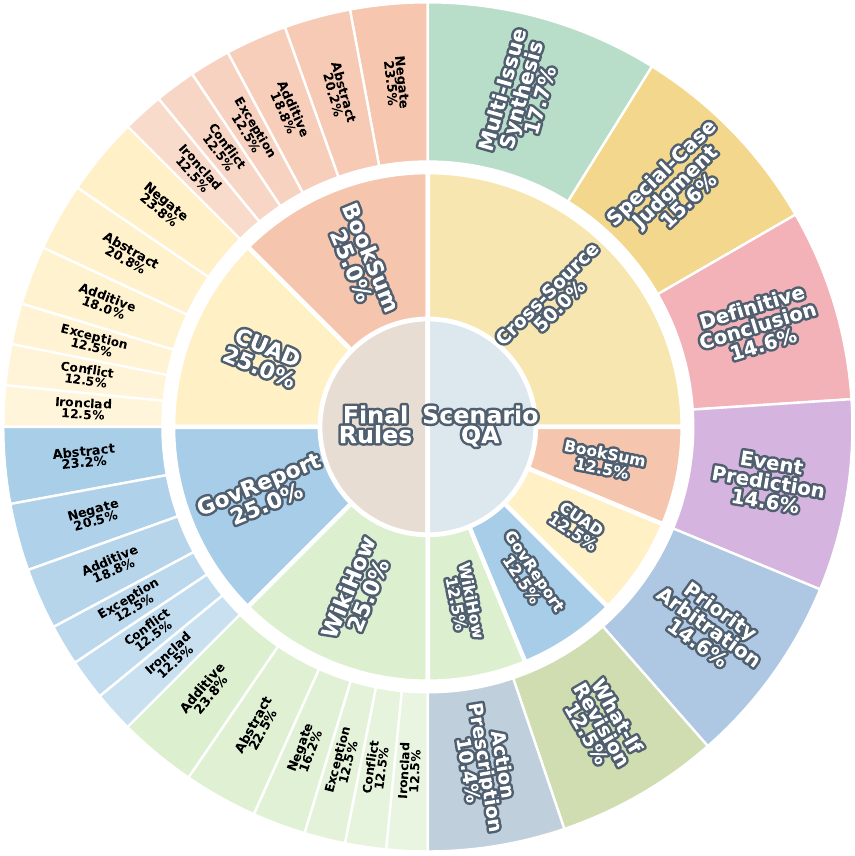}
\end{minipage}
\caption{\textbf{Left:} comparison of RuleWeaver with instruction-following and logical reasoning benchmarks across core evaluation dimensions. \cmark: support; \pmark: partial support; \xmark: unsupported. \textbf{Right:} distribution of the 200 complex rule groups and 96 scenario QA instances. Detailed rule-source and QA-type distributions are provided in Appendices~\ref{app:Final Rule Set Statistics} and~\ref{app:Scenario QA Set Statistics}, respectively, with question-type descriptions in Appendix~\ref{app:question-types}.}
\label{tab:ruleweaver_by_type_time}
\end{table*}

\section{Related Work}
Rule-centered scenario reasoning is closely related to two lines of LLM evaluation: instruction-following benchmarks and logical reasoning benchmarks. Table~\ref{tab:ruleweaver_by_type_time} (left) summarizes representative datasets along these dimensions.

\paragraph{Instruction-Following Benchmarks}
Instruction-following benchmarks mainly evaluate whether models can satisfy constraints embedded in natural language instructions, such as format and style requirements~\cite{jiang-etal-2024-followbench,wizardLM,qin-etal-2024-infobench,zhou2023instructionfollowingevaluationlargelanguage,complex_bench}. These constraints can be viewed as weak rule-like requirements, and some benchmarks further decompose or compose them to evaluate multi-constraint compliance~\cite{jiang-etal-2024-followbench,complex_bench}. However, such benchmarks usually treat rules as output constraints for compliance checking, rather than as reasoning units with different roles and condition-triggered consequences. As a result, they provide limited support for analyzing rule roles, such as abstraction, negation, exception, conflict, and non-overridable constraints, or for evaluating how different types of rules are selected and applied in concrete scenarios.

\paragraph{Logical Reasoning Benchmarks}
Logical reasoning benchmarks~\cite{clark2020ruletaker,tafjord2021proofwritergeneratingimplicationsproofs,saparov2023prontoqa,han2024folionaturallanguagereasoning,parmar2024logicbenchsystematicevaluationlogical,yu2026factualknowledgebenchmarkinglearning} explicitly use rules as premises for inference. Some further introduce formal logic annotations, controllable reasoning depth, or real-world rules~\cite{patel2024multilogievalevaluatingmultisteplogical,zhou2025rulearenabenchmarkruleguidedreasoning,qi2025proverqa}. However, as shown in Table~\ref{tab:ruleweaver_by_type_time} (left), process-level scoring and multi-source or cross-domain rule composition remain underexplored in most existing benchmarks.
More importantly, these benchmarks primarily assess whether models can apply provided rules to derive conclusions, often treating rules as homogeneous inference premises. This makes it difficult to evaluate whether models can distinguish different rule roles, identify rules relevant to a scenario, and combine multiple rules through scenario-specific dependencies. As a result, existing benchmarks do not fully capture the rule selection, role understanding, and dependency-aware application required by rule-centered scenario reasoning.

RuleWeaver bridges these lines by constructing complex rules with different roles from corpus-derived Meta Rules and composing them into concrete QA scenarios. This enables fine-grained evaluation of rule selection, role understanding, dependency-aware application, and process-level answer quality.

\begin{figure*}[htbp]
    \centerline{\includegraphics[scale=0.562]{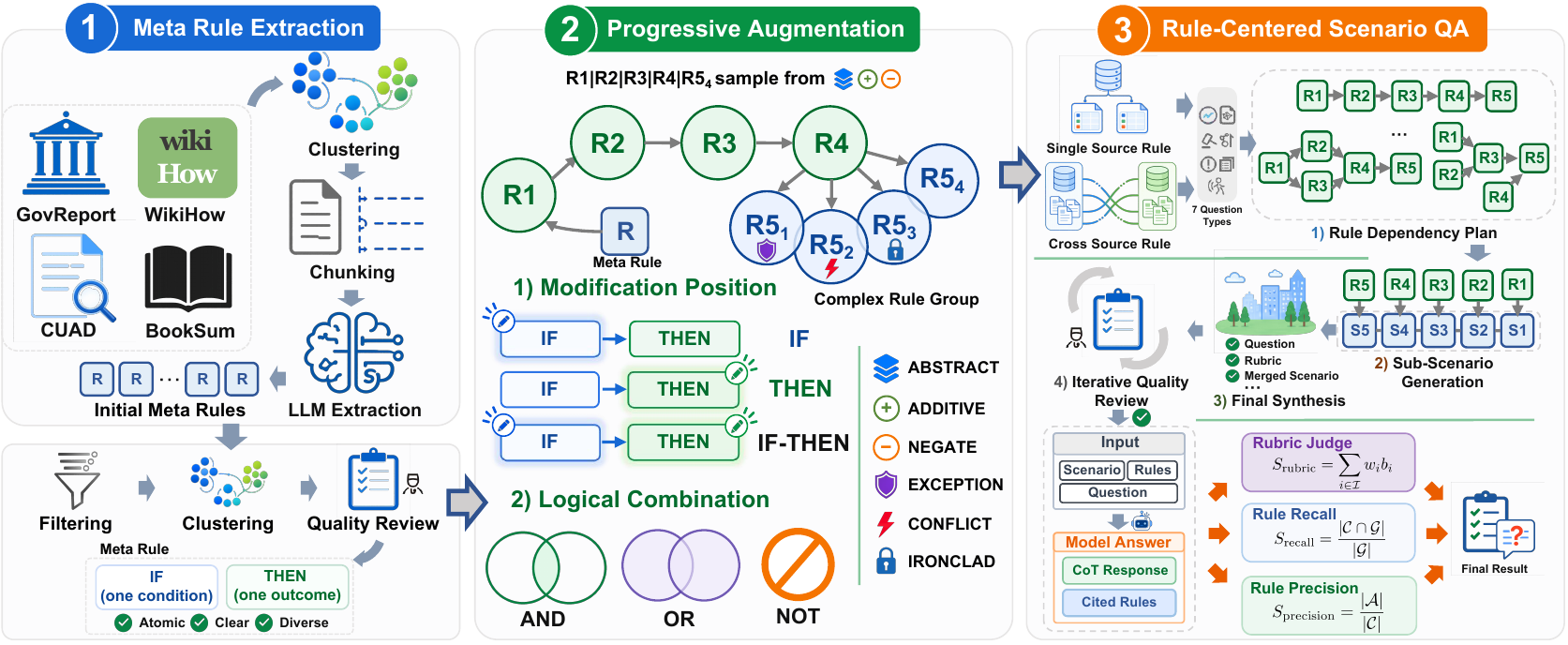}}%
    \caption{Overview of RuleWeaver: from real-world corpora to progressively augmented complex rules, rule-centered scenario QA, and process-level scoring.}
    \label{fig:main}
\end{figure*}

\section{RuleWeaver}
RuleWeaver is a traceable benchmark construction framework for rule-centered scenario reasoning, built from corpus-derived \textit{Meta Rules}. A Meta Rule is an atomic IF-THEN rule with one triggering condition and one outcome, serving as the minimal unit for progressive rule augmentation. As shown in Figure~\ref{fig:main}, RuleWeaver transforms these units into complex rule groups and composes them into rule-centered scenario QA instances.

\subsection{Rule Augmentation}
Rule augmentation defines the semantic design space for transforming a Meta Rule into a more complex variant. We represent each augmentation with three attributes: semantic enhancement type, modification position, and logical combination method. The semantic enhancement type captures the intended rule change and includes six categories: \textbf{ABSTRACT}, \textbf{ADDITIVE}, \textbf{NEGATE}, \textbf{EXCEPTION}, \textbf{CONFLICT}, and \textbf{IRONCLAD}. The first three introduce moderate changes to rule granularity, scope, or polarity, while the latter three model higher-impact rule interactions involving applicability, compatibility, or priority.
The modification position records whether the change applies to the IF side, the THEN side, or both. The logical combination method records how the added information is composed with the original rule through AND, OR, or NOT. Together, these attributes make each augmentation explicit and traceable, enabling controlled variation in rule complexity and semantic type. Detailed definitions and representative examples are provided in Appendix~\ref{app:Representative-Semantic-Augmentation-Examples}.

\subsection{Complex Rule Construction}

We construct complex rule groups in three stages: Meta Rule extraction, quality filtering, and progressive augmentation. Prompts and implementation details for these stages, including generation and filtering models, are provided in Appendix~\ref{app:complex-rule-implementation}.

\paragraph{Initial Meta Rule Construction}

We extract Meta Rules from real corpus documents so that the basic rule units reflect naturally occurring constraint expressions. We use four sources: GovReport~\cite{huang-etal-2021-govreport}, WikiHow~\cite{koupaee2018wikihowlargescaletext}, CUAD~\cite{hendrycks2021cuadexpertannotatednlpdataset}, and BookSum~\cite{kryscinski-etal-2022-booksum}, covering policy reports, procedural guides, contractual clauses, and narrative contexts; examples are shown in Appendix~\ref{app:Representative-Meta-Rule-Examples}.
To improve source coverage, we cluster documents within each corpus and sample representative documents for rule extraction, yielding 11,145 initial Meta Rules.

\paragraph{Meta Rule Quality Filtering}
We filter the extracted rules for format validity, atomicity, semantic clarity, and independence from external background knowledge. To improve diversity and reduce redundancy, we further perform representative clustering and manual inspection, resulting in 200 high-quality Meta Rules. These Meta Rules cover diverse sources and expression styles while preserving clear atomic IF-THEN structure.

\paragraph{Progressive Complex Rule Construction}

Starting from the 200 filtered Meta Rules, we construct complex rules through a five-round progressive augmentation process.
For each Meta Rule, the first four rounds form a shared augmentation trajectory, where one moderate enhancement type is sampled from ABSTRACT, ADDITIVE, and NEGATE and applied to the IF side, THEN side, or both. 
Because EXCEPTION, CONFLICT, and IRONCLAD can substantially affect rule applicability, priority, or rule interactions, they are reserved for the fifth round, where each trajectory branches into four final variants: one sampled from ABSTRACT, ADDITIVE, and NEGATE, and one for each high-impact type.
This yields 200 complex rule groups whose variants collectively cover all six enhancement types, while preserving traceability to the initial Meta Rule, augmentation rounds, modification positions, and logic combinations.

\subsection{Rule-Centered Scenario QA Construction}

After constructing grouped complex rules, we generate rule-centered scenario QA instances through four stages: dependency planning, sub-scenario generation, final synthesis, and quality review. 
We use DeepSeek-V4-Pro~\cite{deepseek2026v4pro} throughout all stages and consider two QA settings: \textit{same-source} QA, where five relevant rules are sampled from a single source dataset, and \textit{cross-source} QA, where the rules are sampled across four datasets.
For each instance, the rule set $\mathcal{R}=\{r_1,\ldots,r_5\}$ is sampled from distinct rule groups to avoid including sibling variants of the same Meta Rule, and each rule is assigned a unique identifier for dependency tracking.
Prompts covering all stages are shown in the Figures~\ref{fig:Dependency_plan_construction_prompt}--~\ref{fig:Scenario_question_answering_quality_judgment_prompt_part_2}.

\paragraph{Dependency Planning}
Given $\mathcal{R}$, the generator first samples a target question type from seven predefined categories, covering multi-issue synthesis, definitive conclusion derivation, event prediction, priority arbitration, special-case judgment, what-if revision, and action prescription (see Appendix~\ref{app:question-types} for detailed definitions).
It then constructs a \textit{rule dependency plan}, a directed structure over rule-application steps that specifies which rule is applied at each step and how intermediate rule-derived conclusions support later steps.
This plan controls the reasoning topology of each QA instance, including independent root rules, sequential dependency transfer, fan-out branching, convergence over multiple rule conclusions, and multi-level aggregation.
By explicitly encoding these dependencies, RuleWeaver can generate scenarios that require models not only to apply individual rules, but also to preserve and combine intermediate conclusions across the rule chain.

\paragraph{Sub-Scenario Generation}
The generator then instantiates each dependency step as a local \textit{sub-scenario}. For each step, it observes the current rule, the already generated predecessor steps, and the next downstream step to which the current step feeds, together with its associated rule. It then writes a scenario fragment together with auxiliary fields used only during construction. These fields include trigger facts, notes on facts that trigger exception or conflict rules, an intermediate conclusion, and the conclusion's downstream contribution. The fragment embeds rule-triggering facts through natural contextual details, while avoiding explicit disclosure of rule text, rule identifiers, hidden sub-questions, or rule conclusions.

\paragraph{Final Synthesis}
Finally, the enriched steps are merged into a single coherent scenario and one final question. The generator also produces the rule-application annotations for computing rule-application accuracy, a \textit{rule logic chain}, and a QA-specific all-or-nothing \textit{rubric}. The rule logic chain serves as the reference reasoning path for rubric-based scoring by recording, for each reasoning step, the rule identifier, dependencies, trigger facts, intermediate conclusion, and downstream contribution. The rubric defines the scoring criteria for each instance, covering aspects such as logical alignment, dependency transfer, exception/conflict handling, and final-answer correctness. 

\paragraph{Quality Review and Refinement}
The generated QA is further refined through an iterative quality-control loop, typically run for five rounds. After a candidate instance is produced, heuristic checks verify basic validity, such as well-formed JSON outputs and coverage of directly used rules. LLM-based judges further assess the dependency plan, sub-scenarios, and final synthesis for structural validity, rule coverage, information leakage, and dependency transfer. The identified issues are summarized into stage-specific feedback and used to optimize the scenario in subsequent refinement rounds. 
Instances whose review scores exceed the predefined quality threshold of 80 points are accepted and subsequently subjected to human inspection.
This loop enables later iterations to produce scenarios with clearer dependencies, better rule coverage, and less information leakage. 

\subsection{Scoring Design}

During evaluation, each model receives a final scenario, a corresponding question, and a visible rule pool containing both relevant and distractor rules, then returns a final answer together with the rule identifiers it cites or applies.
We report three complementary scores: rubric-based answer quality, rule recall, and rule precision.
For answer quality, the LLM judge applies the instance-specific all-or-nothing rubric. The rubric has a total value of 100 points, distributed across its dimensions. Let $\mathcal{I}$ denote the rubric dimensions, let $w_i$ be the point value of dimension $i$ with $\sum_{i\in\mathcal{I}}w_i=100$, and let $b_i\in\{0,1\}$ indicate whether the model answer satisfies dimension $i$. A satisfied dimension receives its full point value; otherwise it receives zero. The rubric score is $S_{\mathrm{rubric}}=\sum_{i\in\mathcal{I}} w_i b_i$.
For rule retrieval, let $\mathcal{G}$ be the set of gold relevant rule identifiers and $\mathcal{C}$ be the set of identifiers cited by the model. Rule recall is $S_{\mathrm{recall}}=\frac{|\mathcal{C}\cap \mathcal{G}|}{|\mathcal{G}|}$.
For rule application, let $\mathcal{A}$ denote the cited rule applications judged correct with respect to the reference answer. Rule precision is $S_{\mathrm{precision}}=\frac{|\mathcal{A}|}{|\mathcal{C}|}$,
with $S_{\mathrm{precision}}$ set to zero when no rule is cited.
Reporting these three scores separately distinguishes answer quality, rule retrieval, and rule application.

\subsection{Human Quality Assurance}

Human quality assurance supplements Rule-\allowbreak{}Weaver's automatic filtering and LLM-based review. Reviewers assess Meta Rules for atomicity, clarity, source faithfulness, and domain relevance; complex rules for traceability, semantic-type correctness, and logical consistency; and scenario QA items for coherent scenarios, questions, rule chains, annotations, and rubrics. Items that are ambiguous, inconsistent, insufficiently rule-grounded, or prone to information leakage are revised or excluded.

\section{Evaluation}
\subsection{Model Selection}
We evaluate 11 state-of-the-art models spanning multiple model families on 96 complex scenario QA instances: 
Claude-Opus-4.6 \cite{anthropic2026claudeopus46systemcard},
Claude-Sonnet-4.6 \cite{anthropic2026claudesonnet46}, 
Deepseek-V4-Pro \cite{deepseek2026v4pro}, 
Qwen3.5-Plus \cite{alibabacloud2026qwen35plus},
GPT-5.4 \cite{openai2026gpt54}, 
GPT-5.5 \cite{openai2026gpt55},
Gemini-3.1-Pro-Preview \cite{google2026gemini31propreview}, 
Doubao-Seed-2.0-pro \cite{bytedance2026seed20pro}, 
GLM-5 \cite{glm5team2026glm5}, 
Kimi-K2.6 \cite{moonshot2026kimik26}, 
and MiniMax-M2.7 \cite{minimax2026m27}. 

\subsection{Evaluation Settings}

For all evaluated models, we disable reasoning modes when configurable and set the decoding temperature to 0 to ensure deterministic generation, except for Kimi-K2.6, whose APIs require the temperature to be set to 0.6 for non-thinking mode. For Gemini-3.1-Pro-Preview, we set the thinking configuration to low.
We use DeepSeek-V4-Flash~\cite{deepseek2026v4pro} as the judge model, with its temperature also set to 0. 
Each instance contains five gold rules, and our main evaluation is conducted under the settings with 200 complex rules.
Rule recall is computed by detecting the rule identifiers explicitly cited in each model response and comparing them with the reference set of required rules. The QA template and judge prompt are provided in Figures~\ref{fig:question_answering_prompt} and~\ref{fig:QA_template_answer_judge}, respectively.

\subsection{Main Results}

\begin{table*}[t]
\centering
\scriptsize
\setlength{\tabcolsep}{4pt}
\renewcommand{\arraystretch}{1.1}
\begin{minipage}[t]{0.75\textwidth}
\vspace{0pt}
\centering
\resizebox{\linewidth}{!}{
\begin{tabular}{lcccccc}
\toprule
\multirow{2}{*}{\textbf{Model}} &
\multicolumn{3}{c}{\textbf{Same-Source}} &
\multicolumn{3}{c}{\textbf{Cross-Source}} \\
\cmidrule(lr){2-4}\cmidrule(lr){5-7}
& \textbf{Recall} & \textbf{Precision} & \textbf{Rubric}
& \textbf{Recall} & \textbf{Precision} & \textbf{Rubric} \\
\midrule
Claude-Opus-4.6 & 70.42 & 70.41 & 48.06 & \textbf{62.50}\drop{7.92} & 73.74\gain{3.33} & \textbf{50.27}\gain{2.21} \\
Claude-Sonnet-4.6 & 71.67 & 63.15 & 43.96 & 59.58\drop{12.09} & 54.74\drop{8.41} & 36.56\drop{7.40} \\
GPT-5.4 & 61.67 & 71.31 & 48.21 & 48.33\drop{13.34} & \textbf{78.64}\gain{7.33} & 42.58\drop{5.63} \\
GPT-5.5 & 69.17 & \textbf{74.58} & \textbf{53.83} & 52.08\drop{17.09} & 69.48\drop{5.10} & 46.40\drop{7.43} \\
Gemini-3.1-Pro-Preview & 60.83 & 58.44 & 31.77 & 45.83\drop{15.00} & 72.95\gain{14.51} & 27.81\drop{3.96} \\
Deepseek-V4-Pro & 63.75 & 59.22 & 34.69 & 55.00\drop{8.75} & 60.10\gain{0.88} & 28.38\drop{6.31} \\
Qwen3.5-Plus & 66.67 & 66.32 & 37.75 & 55.83\drop{10.84} & 64.27\drop{2.05} & 34.15\drop{3.60} \\
GLM-5 & 62.50 & 64.17 & 37.54 & 51.67\drop{10.83} & 69.21\gain{5.04} & 32.35\drop{5.19} \\
Doubao-Seed-2.0-pro & 48.33 & 61.63 & 27.67 & 39.58\drop{8.75} & 75.73\gain{14.10} & 33.02\gain{5.35} \\
MiniMax-M2.7 & 65.00 & 47.43 & 29.69 & 52.08\drop{12.92} & 43.06\drop{4.37} & 21.15\drop{8.54} \\
Kimi-K2.6 & \textbf{72.92} & 55.13 & 32.50 & 59.17\drop{13.75} & 56.53\gain{1.40} & 28.48\drop{4.02} \\
\bottomrule
\end{tabular}
}
\end{minipage}
\begin{minipage}[t]{0.215\textwidth}
\vspace{0pt}
\centering
\includegraphics[width=\linewidth]{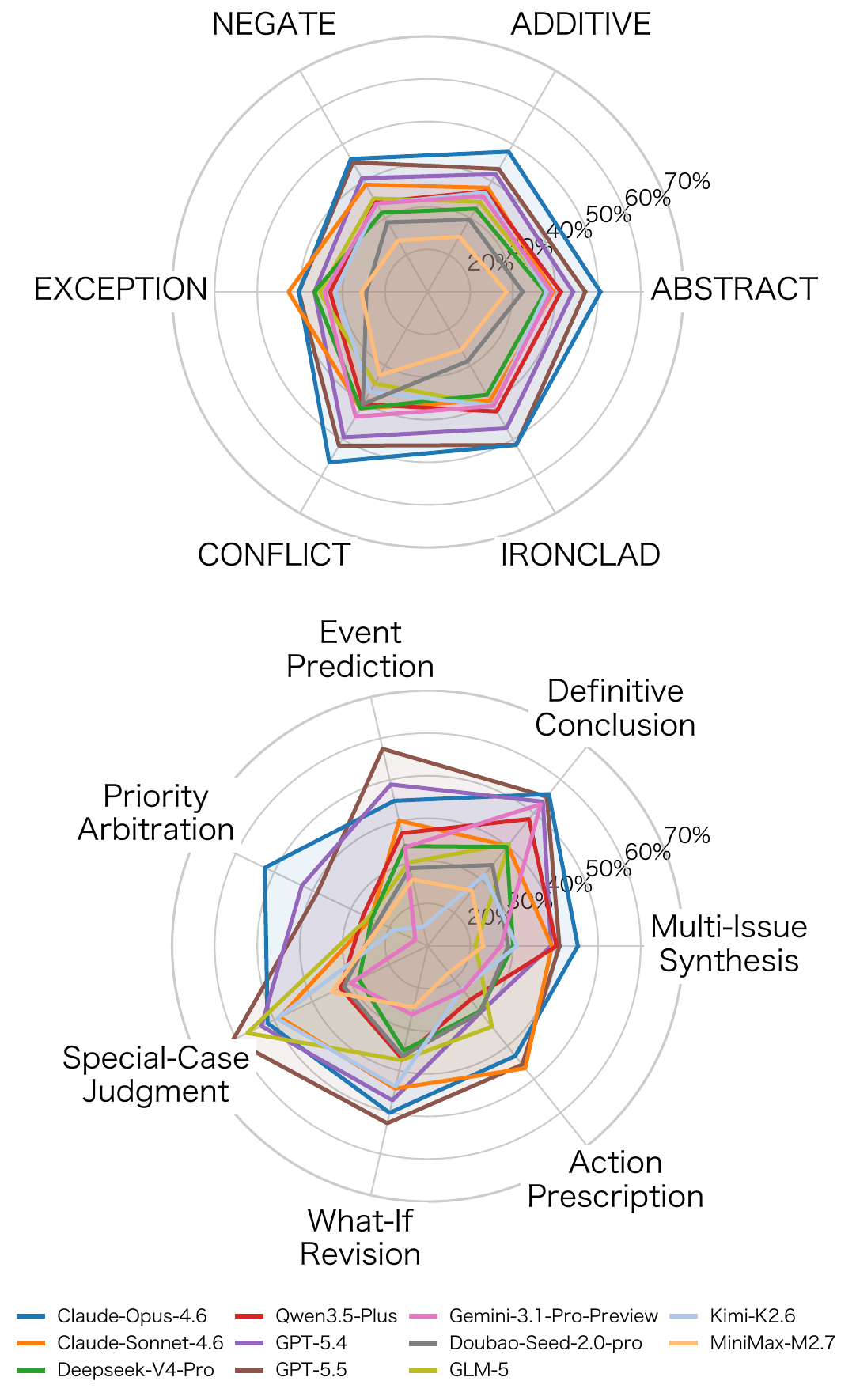}
\end{minipage}
\caption{\textbf{Left:} Main results under the 200 complex-rule setting. Rule recall and precision are percentages, rubric score is on a 0--100 scale, arrows indicate cross-source changes relative to same-source scores, and bold marks the best result in each column. \textbf{Right:} Radar summaries of semantic correct-application performance (top) and seven question-type performance (bottom).}
\label{tab:main_results_200_rules}
\end{table*}

\paragraph{Overall Performance}
Table~\ref{tab:main_results_200_rules} reports the main 200-rule results. Under same-source composition, GPT-5.5 obtains the highest rubric score (53.83) and rule precision (74.58), while Kimi-K2.6 achieves the highest rule recall (72.92). Under cross-source composition, Claude-Opus-4.6 is strongest in both rubric score (50.27) and rule recall (62.50), whereas GPT-5.4 obtains the highest rule precision (78.64). Averaged over the 11 models, cross-source evaluation reduces recall by 11.93 points and rubric score by 4.05 points, while precision increases slightly by 2.42 points. This indicates that heterogeneous rule composition primarily stresses complete rule retrieval and rule-centered scenario reasoning.
Bootstrap analyses further confirm that the reported point estimates are stable under resampling, with detailed results provided in Appendix~\ref{app:bootstrap-analysis}.

\paragraph{Semantic Capabilities}
The upper radar in Table~\ref{tab:main_results_200_rules} analyzes performance by semantic augmentation type, with detailed numerical results provided in Appendix~\ref{app:semantic-results-200}. We score semantics at the enhancement-event level, so a rule may contribute multiple events when its five-round enhancement contains repeated semantic types. 
For a model $m$ and semantic type $s$, the semantic score is the fraction of events in $E(m,s)$ whose parent rule is judged as a correct application, i.e., $\sum_{e\in E(m,s)} \mathbf{1}[\mathrm{rule}(e)\in \mathcal{A}(e)] / |E(m,s)|$, where $\mathcal{A}(e)$ is the set of correctly applied rules in the judge output for the answer containing event $e$. 
Overall, CONFLICT is the strongest semantic category (42.7\%), followed by IRONCLAD (40.6\%) and ABSTRACT (39.7\%), while EXCEPTION is the weakest (34.6\%). 
At the model level, Claude-Opus-4.6 leads five of the six semantic axes, while Claude-Sonnet-4.6 leads EXCEPTION. 
The consistently low EXCEPTION score suggests that models remain especially brittle when rule applicability is conditioned on exception structure rather than direct condition matching.

\paragraph{Performance Across Question Types}
The lower radar summarizes performance across the seven scenario question types under the same 200-rule setting, with detailed question-type definitions and numerical results provided in Appendices~\ref{app:question-types} and~\ref{app:question-type-results-200}. Across models, average rubric scores are highest for Special-Case Judgment (43.5) and Definitive Conclusion (43.4), whereas Priority Arbitration (28.4) and Action Prescription (31.8) are more difficult. The leading model also varies by question type: GPT-5.5 performs best on Event Prediction, Special-Case Judgment, and What-If Revision; Claude-Opus-4.6 leads Multi-Issue Synthesis, Definitive Conclusion, and Priority Arbitration; and Claude-Sonnet-4.6 leads Action Prescription. These differences suggest that model capability is not monolithic, but instead depends on the kind of reasoning a scenario demands.

\paragraph{Rubric-Dimension Performance}
To analyze rubric-level performance loss, we average normalized dimension scores over all model--QA pairs; dimension definitions are provided in Appendix~\ref{app:rubric-dimensions}.
Table~\ref{tab:rubric_dimension_overall_results} shows that models perform well on surface-level constraints, including No External Facts (91.2\%) and Citation Format Compliance (82.8\%).
By contrast, the weakest dimensions are Dependency Chain Alignment (14.2\%), Intermediate Conclusion Quality (21.0\%), Issue Decomposition (28.4\%), and Exception/Conflict Handling (30.4\%).
This pattern suggests that the central bottleneck is maintaining rule-grounded intermediate reasoning across dependent steps, rather than response formatting or unsupported factual additions.
A model-level heatmap is provided in Appendix~\ref{app:rubric-dimension-model-breakdown}. 
Additional QA-level correlation analyses and qualitative case studies are provided in Appendices~\ref{app:qa-level-correlation-results} and~\ref{app:case-studies}.

\begin{table}[t]
\centering
\small
\setlength{\tabcolsep}{4pt}
\renewcommand{\arraystretch}{1.1}
\resizebox{\linewidth}{!}{
\begin{tabular}{lrrr}
\toprule
\textbf{Rubric Dimension} & \textbf{Max} & \textbf{Mean} & \textbf{Norm.} \\
\midrule
Question Understanding & 10 & 5.28 & 52.8\% \\
Issue Decomposition & 10 & 2.84 & 28.4\% \\
Citation Format Compliance & 8 & 6.62 & 82.8\% \\
Rule-Grounded Reasoning & 17 & 6.44 & 37.9\% \\
Dependency Chain Alignment & 18 & 2.55 & 14.2\% \\
Exception/Conflict Handling & 12 & 3.65 & 30.4\% \\
Intermediate Conclusion Quality & 10 & 2.10 & 21.0\% \\
Final Answer Consistency & 12 & 4.36 & 36.3\% \\
No External Facts & 3 & 2.74 & 91.3\% \\
\bottomrule
\end{tabular}
}
\caption{Average rubric-dimension scores under the 200-rule setting. ``Norm.'' denotes the mean score normalized by each dimension's maximum points.}
\label{tab:rubric_dimension_overall_results}
\vspace{-0.5em}
\end{table}

\begin{figure}[t]
    \centering
    \begin{subfigure}[t]{0.95\linewidth}
        \centering
        \includegraphics[width=\linewidth]{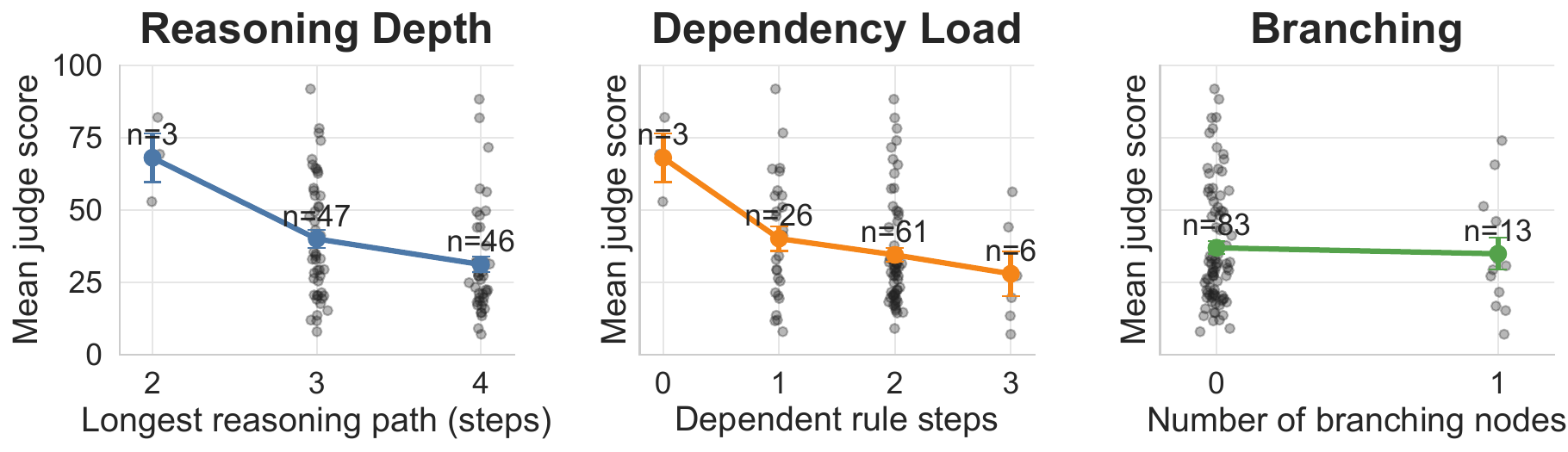}
        \caption{QA-level performance by scenario complexity.}
        \label{fig:complexity_dimension_performance}
    \end{subfigure}

    \begin{subfigure}[t]{0.95\linewidth}
        \centering
        \includegraphics[width=\linewidth]{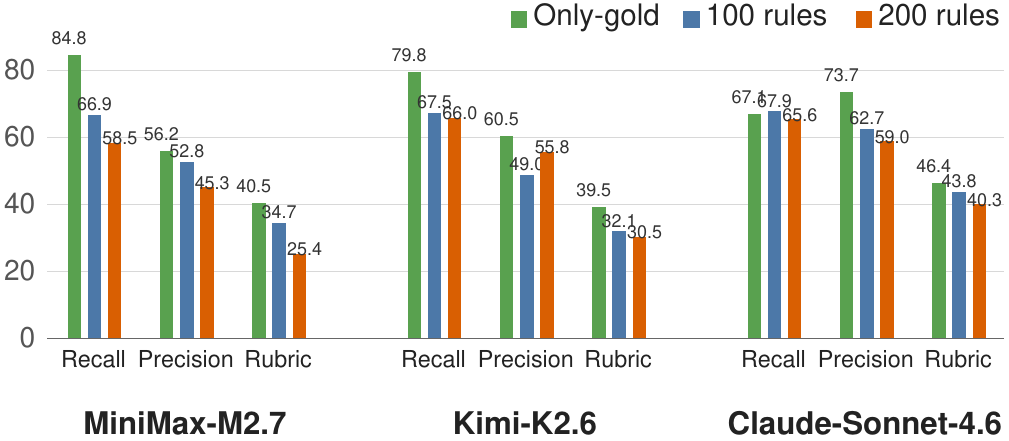}
        \caption{Overall performance under different rule-pool settings.}
        \label{fig:rule_pool_scaling}
    \end{subfigure}

    \caption{\textbf{Top}: gray points show individual QA instances, and colored markers show mean rubric scores with standard-error bars across scenario complexity levels. 
    \textbf{Bottom}: performance under only-gold, 100-rule, and 200-rule settings for three selected models.}
    \label{fig:performance_analysis}
\end{figure}

\subsection{Analysis}

\paragraph{Scenario Complexity Analysis}
\label{sec:scenario-complexity-analysis}

We analyze scenario complexity from each instance's \textit{rule logic chain}, treating reasoning steps as nodes and dependency annotations as directed edges.
Let $\mathcal{V}$ be the set of all reasoning nodes, $\mathcal{V}_{\mathrm{rule}}\subseteq\mathcal{V}$ the rule-application nodes, $\mathcal{P}$ the set of directed paths, and $\deg^{-}(v)$ and $\deg^{+}(v)$ the in- and out-degree of node $v$.
We use three graph metrics: reasoning depth $D=\max_{p\in\mathcal{P}} |p|$, where $|p|$ counts the number of reasoning steps on path $p$; dependency load $L=\sum_{v\in\mathcal{V}_{\mathrm{rule}}}\mathbf{1}[\deg^{-}(v)>0]$, the number of rule steps that depend on earlier conclusions; and structural branching $B=\sum_{v\in\mathcal{V}}\mathbf{1}[\deg^{+}(v)>1]$, the number of nodes feeding multiple downstream steps.
Figure~\ref{fig:complexity_dimension_performance} shows that reasoning depth is the strongest complexity signal.
Mean score decreases from 68.1 at depth 2 to 40.0 at depth 3 and 31.2 at depth 4, indicating that longer rule chains are substantially harder for current models.
Dependency load shows a similar trend: scores decline from 68.1 with no dependent rule step to 28.0 with three dependent rule steps.
In contrast, structural branching has little visible association with performance, with branch-free and one-branch scenarios showing similar mean scores.
Overall, sequential depth and the need to reuse intermediate conclusions emerge as the main structural difficulty factors in the current benchmark.

\paragraph{Rule-Pool Size Sensitivity}

To assess sensitivity to rule-pool size, we evaluate three settings: only-gold rules, 100 rules, and 200 rules. Figure~\ref{fig:rule_pool_scaling} reports recall, precision, and rubric score for MiniMax-M2.7, Kimi-K2.6, and Claude-Sonnet-4.6. Performance generally declines as the rule pool expands. MiniMax-M2.7 shows the clearest degradation, with recall dropping from 84.8 to 58.5, precision from 56.2 to 45.3, and rubric score from 40.5 to 25.4. Kimi-K2.6 also drops substantially in recall and rubric score, although its precision partially recovers from the 100-rule setting while remaining below only-gold. Overall, larger rule pools mainly increase the distractor burden: models must identify the relevant rules among more candidates, and final score further depends on whether the selected rules are correctly applied and integrated into the reasoning chain.

\begin{table}[t]
\centering
\small
\setlength{\tabcolsep}{4pt}
\renewcommand{\arraystretch}{1.1}
\begin{tabular}{lrrrr}
\toprule
\multirow{2}{*}{\textbf{Error Type}} & \multicolumn{2}{c}{\textbf{All}} & \multicolumn{2}{c}{\textbf{Full Recall}} \\
\cmidrule(lr){2-3}\cmidrule(lr){4-5}
 & \textbf{Count} & \textbf{Rate} & \textbf{Count} & \textbf{Rate} \\
\midrule
Rule selection & 762 & 72.2\% & -- & -- \\
Cited-rule application & 580 & 54.9\% & 191 & 65.6\% \\
Rule interaction & 734 & 69.5\% & 166 & 57.0\% \\
Multi-step integration & 916 & 86.7\% & 218 & 74.9\% \\
Final synthesis & 678 & 64.2\% & 193 & 66.3\% \\
\bottomrule
\end{tabular}
\caption{Error-type prevalence over all model--QA answers and the subset with full rule recall.}
\label{tab:error_type_prevalence}
\end{table}

\paragraph{Error Type Analysis}
To complement rubric-dimension scores, we group failures into five rule-centered error types using existing evaluation annotations. Rule selection errors are identified from missed gold rules in rule recall; cited-rule application errors from rule precision annotations; rule interaction errors from exception/conflict handling annotations; multi-step integration errors from issue decomposition, intermediate-conclusion, and dependency-chain annotations; and final synthesis errors from final-answer consistency annotations. Because an answer can fail in multiple ways, the analysis is multi-label.
As shown in Table~\ref{tab:error_type_prevalence}, the most frequent error is multi-step integration (86.7\%), followed by rule selection (72.2\%) and rule interaction (69.5\%). In the full-recall subset, multi-step integration errors (74.9\%) and cited-rule application errors (65.6\%) remain frequent, suggesting that models often fail not only by missing relevant rules, but also by misusing retrieved rules or failing to connect intermediate conclusions.

\paragraph{Human-Judge Agreement}
\label{sec:human-judge-agreement}
To assess the reliability of rubric-based automatic evaluation, we compare LLM-judge total scores with human judgments on score-stratified samples. Detailed sampling and annotation procedures are provided in Appendix~\ref{app:human-judge-agreement}. Table~\ref{tab:human_judge_agreement} reports agreement on total rubric scores. The GPT-5.5 audit yields Spearman's $\rho=0.808$~\cite{spearman1904proof} and Pearson's $r=0.795$~\cite{Pearson} between human and LLM-judge scores. Across the four additional audits, judge--human Spearman correlations range from 0.755 to 0.842 and Pearson correlations from 0.635 to 0.826; human--human Spearman correlations range from 0.935 to 0.977 and Pearson correlations from 0.901 to 0.969. These results indicate strong agreement between human and automatic scoring, supporting the reliability of rubric-based evaluation.

\begin{table*}[t]
\centering
\scriptsize
\setlength{\tabcolsep}{2pt}
\renewcommand{\arraystretch}{1.1}
\resizebox{\textwidth}{!}{%
\begin{tabular}{c@{\hspace{2em}}lcccc}
\toprule
\multicolumn{1}{c}{\textbf{GPT-5.5 Audit}} & \multicolumn{5}{c}{\textbf{Additional Cross-Model Audits}} \\
\cmidrule(r){1-1}\cmidrule(l){2-6}
\multirow{5}{*}{\begin{tabular}{@{}c@{}}
\textbf{Judge--human agreement} \\
Spearman's $\rho$: 0.808 \\
Pearson's $r$: 0.795
\end{tabular}}
& \textbf{Audit outputs} & \textbf{Judge--human $\rho$} & \textbf{Judge--human $r$} & \textbf{Human--human $\rho$} & \textbf{Human--human $r$} \\
& Claude-Opus-4.6 & 0.755 & 0.733 & 0.977 & 0.969 \\
& Claude-Sonnet-4.6 & 0.842 & 0.826 & 0.945 & 0.952 \\
& DeepSeek-V4-Pro & 0.795 & 0.635 & 0.947 & 0.901 \\
& GPT-5.4 & 0.834 & 0.762 & 0.935 & 0.902 \\
\bottomrule
\end{tabular}
}
\caption{Agreement on total rubric scores. Each audit contains 48 answers. $\rho$: Spearman; $r$: Pearson.}
\label{tab:human_judge_agreement}
\end{table*}

\paragraph{Evaluator Robustness Analysis}
\label{sec:evaluator-robustness}
To assess possible model-family bias in automatic scoring, we rescore the same outputs with Grok-4.3~\cite{xai2026grok43} and Gemini-3-Flash~\cite{google2025gemini3flash} and compare their model-level mean rubric scores with the DeepSeek-V4-Flash scores. Both alternative judges use temperature 0; Grok-4.3 uses \texttt{reasoning\_effort=none}, and Gemini-3-Flash uses low reasoning effort. In Table~\ref{tab:judge_ablation}, Overall denotes correlations computed from each model's mean rubric score across both same-source and cross-source instances. The high correlations across both source settings and Overall indicate that the main leaderboard patterns are consistent under judges from different model families.

\begin{table}[t]
\centering
\small
\setlength{\tabcolsep}{4pt}
\renewcommand{\arraystretch}{1.1}
\begin{tabular}{llcc}
\toprule
\textbf{Judge} & \textbf{Setting} & \textbf{Spearman $\rho$} & \textbf{Pearson $r$} \\
\midrule
\multirow{3}{*}{Grok-4.3} & Same-source & 0.882 & 0.878 \\
 & Cross-source & 0.918 & 0.860 \\
 & Overall & 0.918 & 0.893 \\
\multirow{3}{*}{Gemini-3-Flash} & Same-source & 0.845 & 0.818 \\
 & Cross-source & 0.873 & 0.793 \\
 & Overall & 0.882 & 0.846 \\
\bottomrule
\end{tabular}
\caption{Correlation of alternative-judge and DeepSeek-V4-Flash model-level mean rubric scores.}
\label{tab:judge_ablation}
\end{table}

\section{Conclusion}
This paper presents \textbf{RuleWeaver}, a benchmark construction framework that transforms corpus-derived Meta Rules into complex rule-centered scenario QA instances. Beyond final-answer correctness, RuleWeaver evaluates models through rubric-based answer quality, rule recall, and rule precision. Experiments on 11 LLMs show that current models remain limited in complex rule-centered scenario reasoning, especially under cross-source composition, exception and priority reasoning, longer reasoning chains, and larger rule pools. These findings highlight rule-centered scenario reasoning as a key challenge for current LLMs.

\section*{Limitations}
RuleWeaver remains a controlled benchmark. Its 96 scenario QA instances are built from four English corpora, leaving broader multilingual, domain-specific, and institutional rule settings for future expansion. The current evaluation uses an explicit rule pool and static single-turn QA format, whereas real applications may require retrieval from long documents, interaction with users, or updates to changing rule sets. Future work should extend RuleWeaver to larger rule collections, unstructured-document retrieval, interactive settings, multilingual scenarios, and alternative judging protocols.
These extensions could further examine structured and transferable memory for long-horizon interactive reasoning~\cite{liang2026learningremembermetacognitivemanagement}, cross-lingual consistency of rule representations~\cite{cao2024mindtonguesdeepdive}, representation-level adaptation for multistep reasoning~\cite{liang2025biasrestrainedprefixrepresentationfinetuning}, and reasoning-consistent updates to changing, interdependent rule sets~\cite{Zhang2026TowardsPK}.

\bibliography{custom}

\appendix
\label{sec:appendix}

\begin{table*}[t]
\centering
\scriptsize
\setlength{\tabcolsep}{3.5pt}
\renewcommand{\arraystretch}{1.18}
\begin{tabular}{llrrrrrrrr}
\toprule
\textbf{Source Corpus} & \textbf{Rule Domain} & \textbf{Groups} & \textbf{Variants} & \textbf{ABSTRACT} & \textbf{ADDITIVE} & \textbf{NEGATE} & \textbf{EXCEPTION} & \textbf{CONFLICT} & \textbf{IRONCLAD} \\
\midrule
GovReport & Policy reports & 50 & 200 & 19 & 16 & 15 & 50 & 50 & 50 \\
WikiHow & Procedural guides & 50 & 200 & 21 & 16 & 13 & 50 & 50 & 50 \\
CUAD & Contractual clauses & 50 & 200 & 15 & 16 & 19 & 50 & 50 & 50 \\
BookSum & Narrative contexts & 50 & 200 & 23 & 15 & 12 & 50 & 50 & 50 \\
\midrule
\textbf{Total} & -- & \textbf{200} & \textbf{800} & \textbf{78} & \textbf{63} & \textbf{59} & \textbf{200} & \textbf{200} & \textbf{200} \\
\bottomrule
\end{tabular}
\caption{Statistics of the complex final rule set.}
\label{tab:organized_rule_statistics}
\end{table*}

\begin{table*}[t]
\centering
\scriptsize
\setlength{\tabcolsep}{2.6pt}
\renewcommand{\arraystretch}{1.16}
\begin{minipage}[t]{0.66\textwidth}
\centering
\textbf{(a) QA source, rule-reference, and scenario-length statistics}\\[2pt]
\begin{tabular}{@{}llrrrrrl@{}}
\toprule
\textbf{QA Source} & \textbf{Mode} & \textbf{QAs} & \shortstack{\textbf{Rule}\\\textbf{Mentions}} & \shortstack{\textbf{Avg.}\\\textbf{Rules}} & \shortstack{\textbf{Avg.}\\\textbf{Words}} & \textbf{Median} & \shortstack{\textbf{Word}\\\textbf{Range}} \\
\midrule
Cross-Source & Cross-source & 48 & 240 & 5.0 & 705.1 & 622.5 & 241--1,425 \\
BookSum & Same-source & 12 & 60 & 5.0 & 796.8 & 669.5 & 349--1,499 \\
CUAD & Same-source & 12 & 60 & 5.0 & 598.8 & 584.5 & 154--1,391 \\
GovReport & Same-source & 12 & 60 & 5.0 & 912.1 & 994.0 & 289--1,262 \\
WikiHow & Same-source & 12 & 60 & 5.0 & 896.0 & 1,022.0 & 307--1,476 \\
\midrule
\textbf{Total} & -- & \textbf{96} & \textbf{480} & \textbf{5.0} & \textbf{753.0} & \textbf{656.5} & \textbf{154--1,499} \\
\bottomrule
\end{tabular}
\end{minipage}
\hfill
\begin{minipage}[t]{0.31\textwidth}
\centering
\textbf{(b) Question type distribution}\\[2pt]
\begin{tabular}{@{}lrr@{}}
\toprule
\textbf{Question Type} & \textbf{QAs} & \textbf{Percent} \\
\midrule
Multi-Issue Synthesis & 17 & 17.71\% \\
Special-Case Judgment & 15 & 15.62\% \\
Definitive Conclusion & 14 & 14.58\% \\
Event Prediction & 14 & 14.58\% \\
Priority Arbitration & 14 & 14.58\% \\
What-If Revision & 12 & 12.50\% \\
Action Prescription & 10 & 10.42\% \\
\midrule
\textbf{Total} & \textbf{96} & \textbf{100.00\%} \\
\bottomrule
\end{tabular}
\end{minipage}
\caption{Statistics of the scenario QA set, including source balance, rule references, scenario length, and question types. Each scenario QA instance is grounded in five complex rules.}
\label{tab:scenario_qa_statistics}
\end{table*}

\section{Data Statistics}

\subsection{Final Rule Set Statistics}
\label{app:Final Rule Set Statistics}

Table~\ref{tab:organized_rule_statistics} summarizes the organized final rule set used by RuleWeaver.
The set contains 200 complex rule groups, corresponding to the 200 filtered Meta Rules selected after corpus-specific clustering and manual inspection.
The groups are balanced across four source corpora, with 50 groups each from GovReport, WikiHow, CUAD, and BookSum, covering policy reports, procedural guides, contractual clauses, and narrative contexts.
Following the progressive construction procedure, each group contains four final variants: one moderate continuation sampled from \textbf{ABSTRACT}, \textbf{ADDITIVE}, and \textbf{NEGATE}, and one variant for each high-impact type, \textbf{EXCEPTION}, \textbf{CONFLICT}, and \textbf{IRONCLAD}.
This yields 800 final complex rules in total.
The high-impact types are exactly balanced by construction, while the moderate continuation branch contains 78 \textbf{ABSTRACT}, 63 \textbf{ADDITIVE}, and 59 \textbf{NEGATE} variants.
The root Meta Rules contain 18.3 words on average, whereas the final complex rule variants contain 71.4 words on average, reflecting the intended accumulation of conditions, outcomes, and semantic interactions.

\subsection{Scenario QA Set Statistics}
\label{app:Scenario QA Set Statistics}

Table~\ref{tab:scenario_qa_statistics} summarizes the scenario QA instances generated from the final rule set.
The current scenario QA set contains 96 instances, each grounded in exactly five complex rules, yielding 480 rule mentions in total.
The data are balanced between 48 cross-source instances and 48 same-source instances.
For same-source QA, the four source corpora each contribute 12 instances; for cross-source QA, each instance samples rules across the four corpora, producing a near-balanced rule-source mixture with 59 BookSum, 64 CUAD, 55 GovReport, and 62 WikiHow rule mentions.
The final scenarios contain 72,291 words in total, with an average length of 753.0 words per instance (median 656.5; range 154--1,499).
Average scenario length ranges from 598.8 words for CUAD-sourced QA to 912.1 words for GovReport-sourced QA.
The question types are also broadly distributed: the most frequent type, \textit{Multi-Issue Synthesis}, accounts for 17 instances, while the remaining six types range from 10 to 15 instances.
Across all QA instances, the 480 referenced rules cover all six semantic augmentation types: 86 \textbf{ABSTRACT}, 82 \textbf{ADDITIVE}, 75 \textbf{NEGATE}, 82 \textbf{EXCEPTION}, 89 \textbf{CONFLICT}, and 66 \textbf{IRONCLAD} rule mentions.

\begin{table*}[t]
\centering
\small
\setlength{\tabcolsep}{3.5pt}
\renewcommand{\arraystretch}{2.1}
\begin{tabular}{p{0.10\textwidth} p{0.21\textwidth} p{0.28\textwidth} p{0.32\textwidth}}
\toprule
\textbf{Type} & \textbf{Source Meta Rule} & \textbf{Augmented Information} & \textbf{Augmentation Definition} \\
\midrule

\textbf{ABSTRACT} &
If a third-party claim arises from an intellectual property infringement, then the indemnifying party should defend the indemnified party. &
The condition is expanded with the broader alternative ``or any other similar legal wrong.'' &
This abstracts the concrete trigger, \textit{intellectual property infringement}, into a higher-level legal category, requiring models to match the rule beyond surface lexical overlap. \\

\textbf{ADDITIVE} &
If an item is controlled on the applicable export control list, then the exporter should submit a license application to the relevant regulatory authority. &
Added condition: ``the item is not explicitly pre-approved for export under a special government-to-government agreement.'' Added outcome: ``not proceed with the export without first obtaining a binding advisory opinion.'' &
The new information does not reverse or suspend the export-control rule; it supplements the original scope with an additional eligibility condition and an additional compliance requirement. \\

\textbf{NEGATE} &
If a site scores above 28.5 on the applicable hazard assessment metric, then the site is eligible for placement on the relevant national priority list. &
The rule becomes: if the site ``does not score above 28.5,'' then it is ``not eligible for placement.'' &
Both the IF condition and THEN outcome are reversed, testing whether a model tracks the changed semantic direction instead of applying the original eligibility rule. \\

\textbf{EXCEPTION} &
If a party wishes to assign the Agreement, then that party should obtain prior written consent from the other party. &
Added exception: ``except where the assignment is to an affiliate.'' &
The added affiliate case restricts the general consent requirement by specifying a special circumstance in which the rule should not be directly applied. \\

\textbf{CONFLICT} &
If the service provider uses commercially reasonable efforts to correct a Program Error, then failure to correct it within the specified period shall not be deemed a breach. &
Added competing condition: if the failure results from ``willful misconduct or fraud,'' then the failure ``shall be deemed a breach'' and triggers penalties. &
The added rule creates an incompatible outcome with the original no-breach conclusion, requiring conflict detection and resolution when both conditions are relevant. \\

\textbf{IRONCLAD} &
If a federal agency intends to dispose of property, then the agency should assess the need for environmental cleanup. &
Added mandatory prohibition: the agency ``must not transfer the property to any non-federal entity before the assessment is complete.'' &
The new item is framed as a strict, non-optional constraint that must be preserved even when other outcomes, waivers, or procedural alternatives are considered. \\

\bottomrule
\end{tabular}
\caption{Representative examples of the six semantic augmentation types selected from the organized final rule set.}
\label{tab:semantic_augmentation_examples}
\end{table*}

\section{Additional Details of RuleWeaver}
\label{app:ruleweaver-details}

\begin{table*}[t]
\centering
\small
\setlength{\tabcolsep}{4pt}
\renewcommand{\arraystretch}{1.25}
\begin{tabular}{p{0.10\textwidth} p{0.36\textwidth} p{0.24\textwidth} p{0.24\textwidth}}
\toprule
\textbf{Source} & \textbf{Meta Rule} & \textbf{IF Condition} & \textbf{THEN Outcome} \\
\midrule

BookSum & 
If a person is excommunicated from the Church, then they are banished from entering Rome. & 
A person is excommunicated from the Church & 
They are banished from entering Rome \\

CUAD & 
If the manufacturer intends to change a manufacturing process, then the manufacturer should obtain written approval from the client via a formal change order. & 
The manufacturer intends to change a manufacturing process & 
The manufacturer should obtain written approval from the client via a formal change order \\

GovReport & 
If the postal service provider proposes to eliminate a standard delivery day, then the provider should request an advisory opinion from the relevant regulatory oversight body. & 
The postal service provider proposes to eliminate a standard delivery day & 
The provider should request an advisory opinion from the relevant regulatory oversight body \\

WikiHow & 
If a tank lacks a bubbler, then the water becomes acidic. & 
A tank lacks a bubbler & 
The water becomes acidic \\

\bottomrule
\end{tabular}
\caption{Examples of Meta Rules from Four Source Datasets.}
\label{tab:meta_rule_examples}
\end{table*}

\subsection{Representative Semantic Augmentation Examples}
\label{app:Representative-Semantic-Augmentation-Examples}

Table~\ref{tab:semantic_augmentation_examples} presents representative examples selected from the organized final rule set.
For readability, the table reports the source Meta Rule and the specific augmented information that carries the semantic type, rather than printing the full five-round complex rule.
These examples illustrate how the same IF-THEN backbone can be enriched in different semantic directions while preserving an explicit record of the augmentation operator and its intended reasoning effect.
The selected examples emphasize that the six augmentation types differ not only in surface wording, but also in the reasoning behavior they induce.
\textbf{ABSTRACT} changes the granularity of a rule term, \textbf{ADDITIVE} integrates further requirements within the same rule scope, and \textbf{NEGATE} changes the direction of rule applicability or consequence.
The higher-impact types operate over rule applicability and priority: \textbf{EXCEPTION} carves out a special non-applicable case, \textbf{CONFLICT} introduces a competing rule with an incompatible conclusion, and \textbf{IRONCLAD} adds a constraint that should remain active despite other rule interactions.

\begin{figure*}[t]
    \centering
    \begin{subfigure}[t]{0.32\textwidth}
        \centering
        \includegraphics[width=\linewidth]{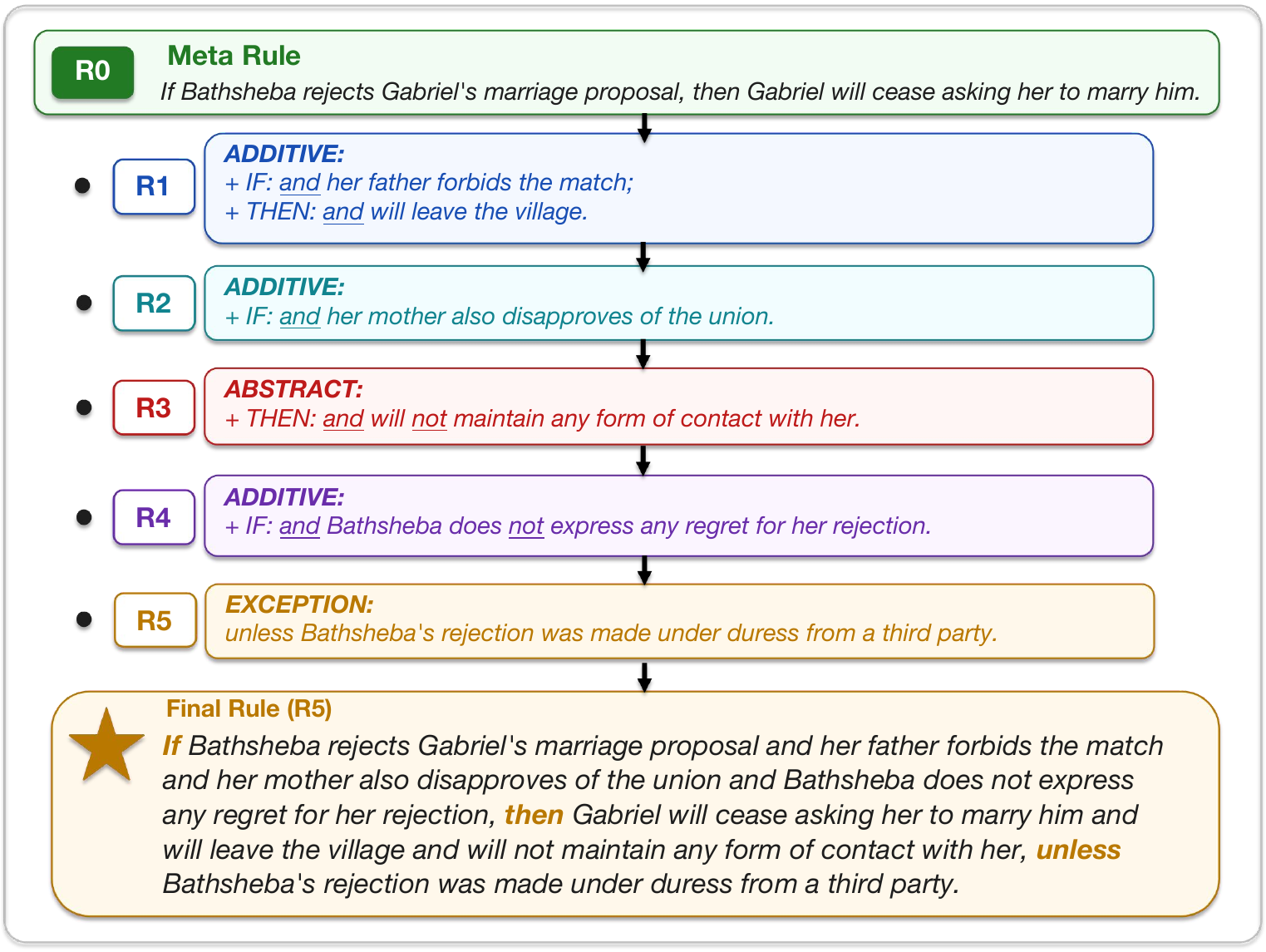}
        \caption{\textbf{EXCEPTION}}
        \label{fig:additional_progressive_exception}
    \end{subfigure}\hfill
    \begin{subfigure}[t]{0.32\textwidth}
        \centering
        \includegraphics[width=\linewidth]{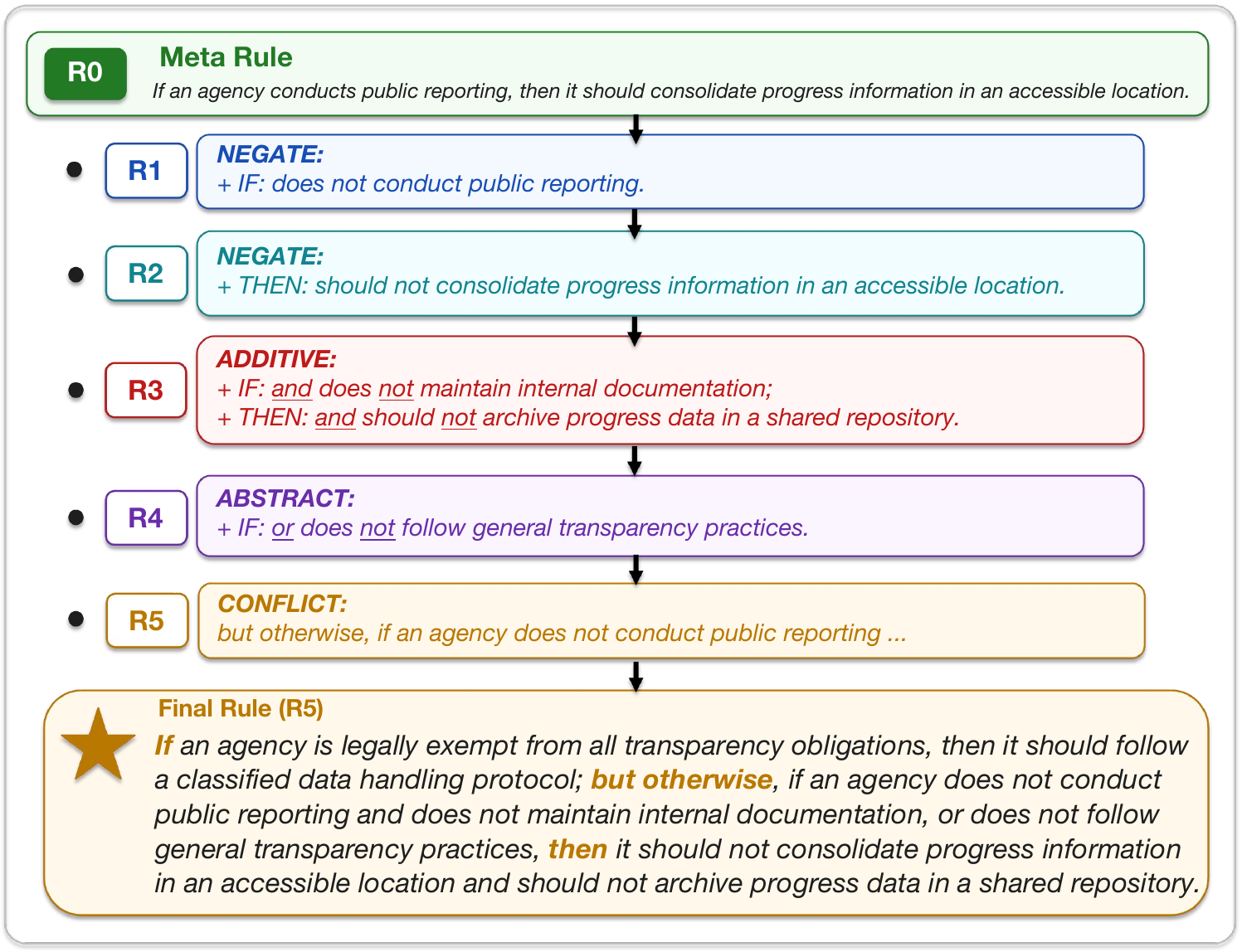}
        \caption{\textbf{CONFLICT}}
        \label{fig:additional_progressive_conflict}
    \end{subfigure}\hfill
    \begin{subfigure}[t]{0.32\textwidth}
        \centering
        \includegraphics[width=\linewidth]{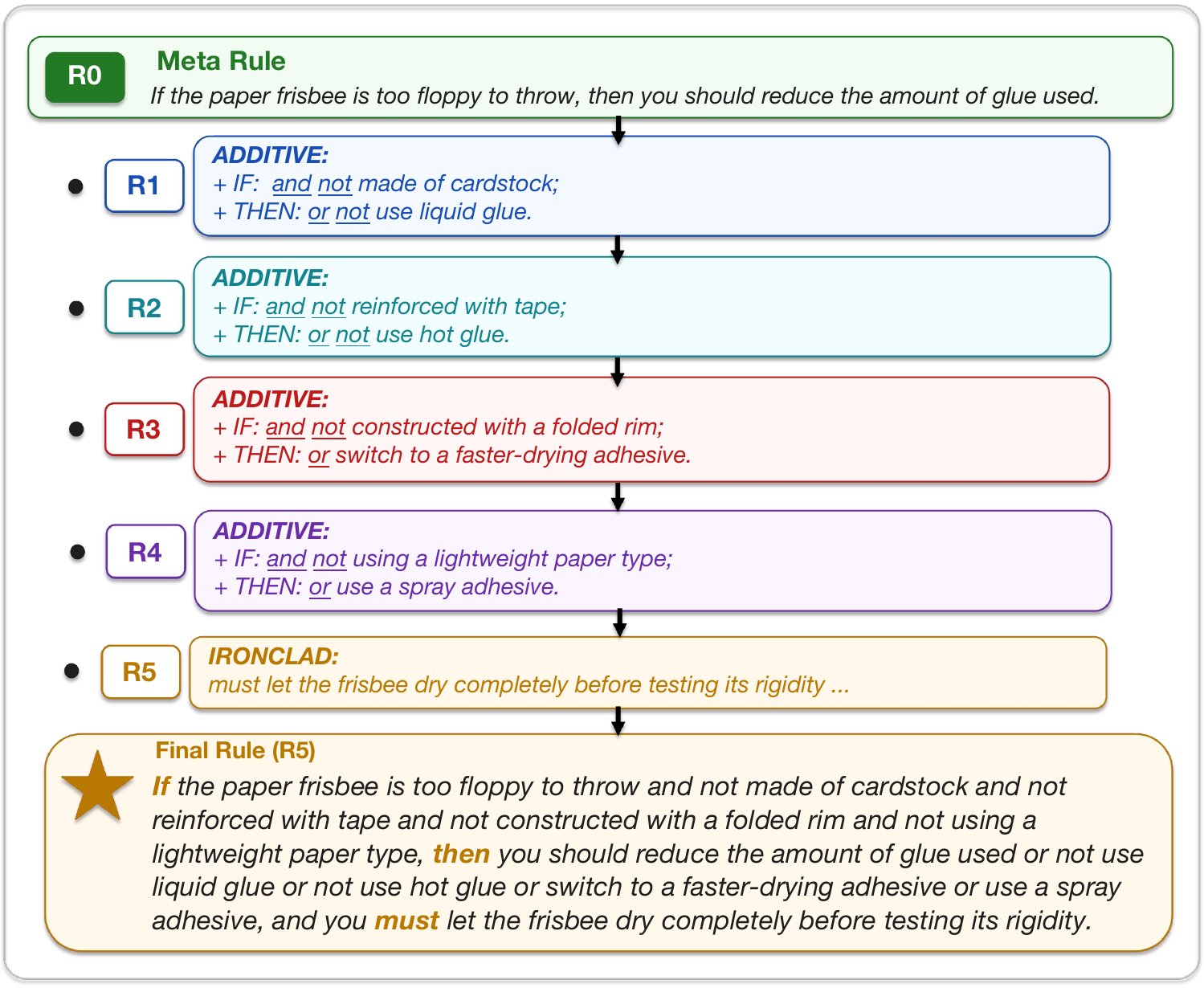}
        \caption{\textbf{IRONCLAD}}
        \label{fig:additional_progressive_ironclad}
    \end{subfigure}
    \caption{Complete five-round progressive augmentation cases, with final semantic enhancements of \textbf{EXCEPTION}, \textbf{CONFLICT}, and \textbf{IRONCLAD}, respectively.}
    \label{fig:additional_progressive_cases}
\end{figure*}

The table above illustrates isolated, single-step semantic augmentation events. To complement these examples, Figure~\ref{fig:additional_progressive_cases} shows three complete five-round progressive augmentation cases. Each case preserves its full augmentation trajectory from the source Meta Rule to the final complex rule: the final round realizes an \textbf{EXCEPTION} in Figure~\ref{fig:additional_progressive_exception}, a \textbf{CONFLICT} in Figure~\ref{fig:additional_progressive_conflict}, and an \textbf{IRONCLAD} constraint in Figure~\ref{fig:additional_progressive_ironclad}. These cases make explicit how earlier moderate augmentations are accumulated before a high-impact rule interaction is introduced in the final round.

\subsection{Implementation Details for Complex Rule Construction}
\label{app:complex-rule-implementation}

\begin{figure}[htbp]
    \centering
    \includegraphics[scale=0.23]{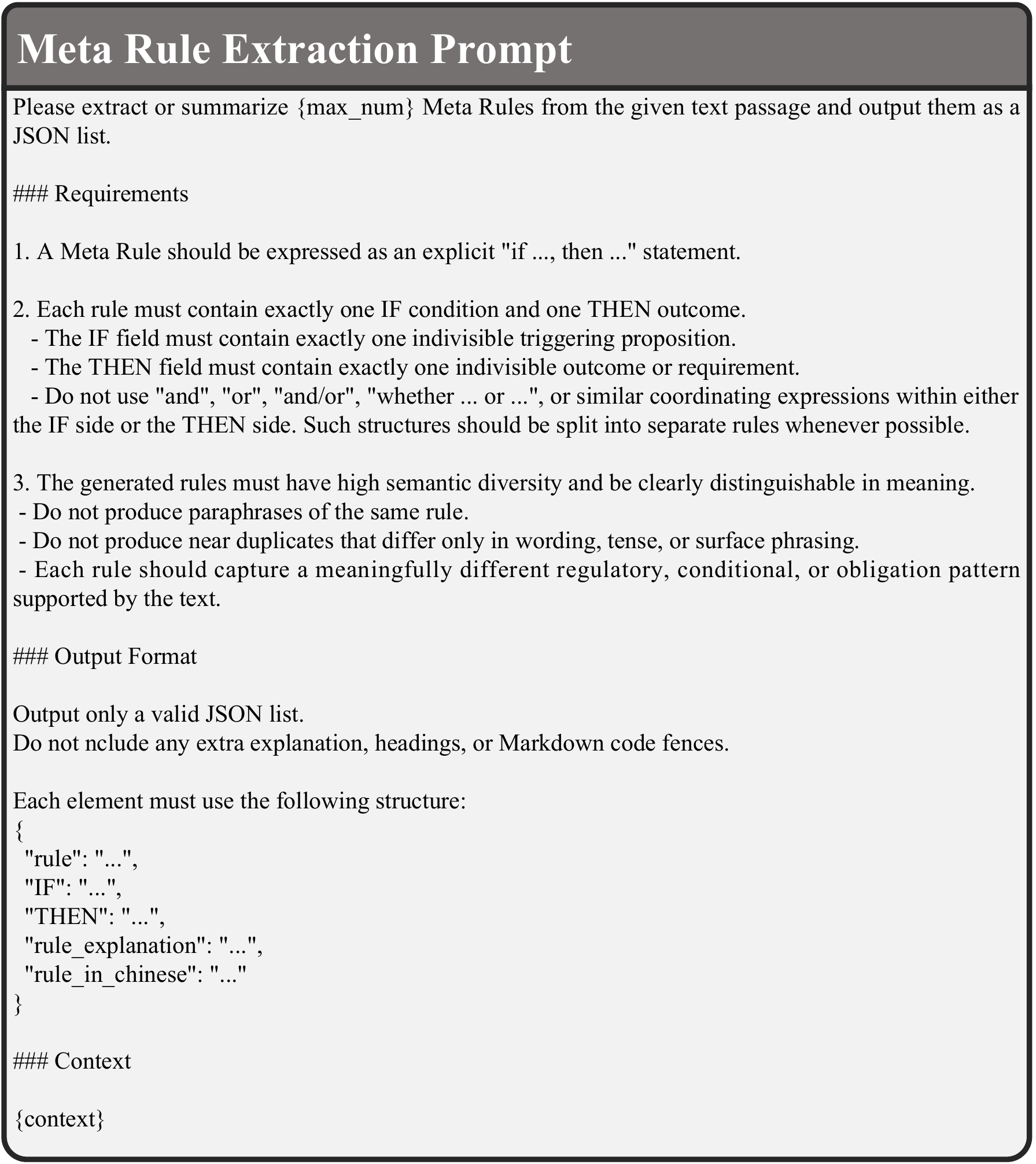}
    \caption{Meta rule extraction prompt.}
    \label{fig:Meta_rule_extraction_prompt}
\end{figure}

\begin{figure}[htbp]
    \centering
    \includegraphics[scale=0.23]{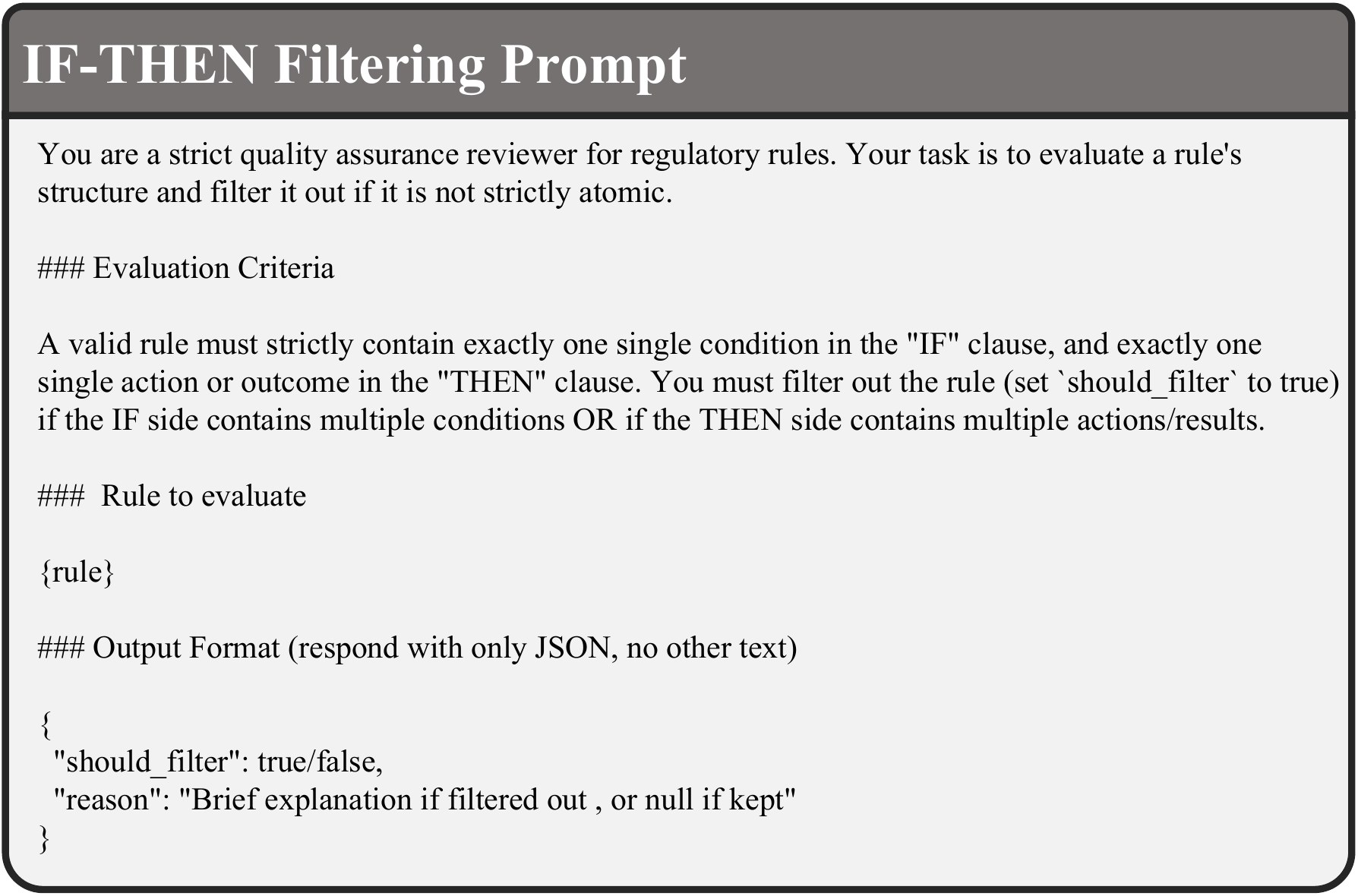}
    \caption{IF-THEN filtering prompt.}
    \label{fig:if_then_filter_prompt}
\end{figure}

\begin{figure}[htbp]
    \centering
    \includegraphics[scale=0.23]{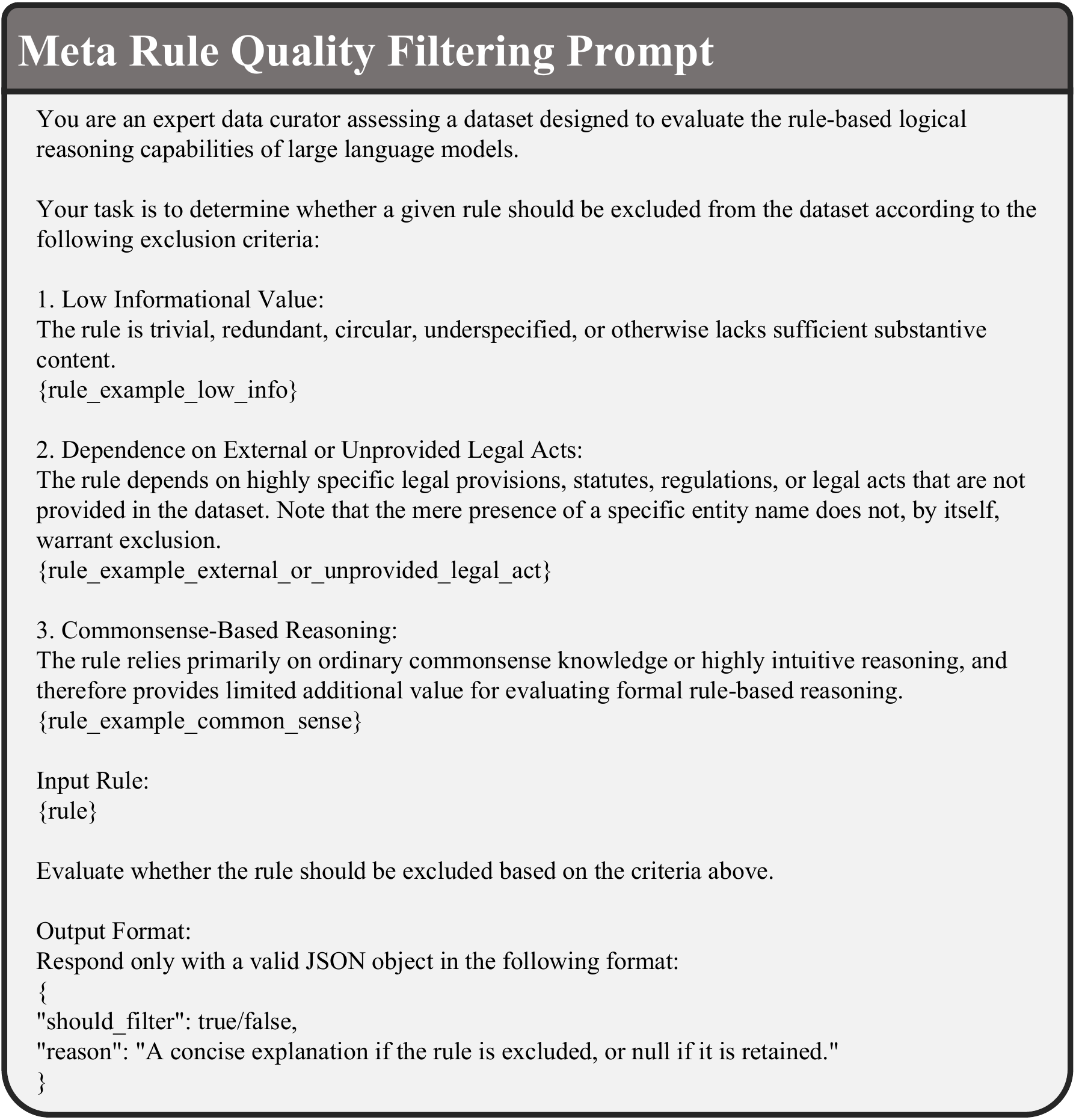}
    \caption{Meta rule quality filtering prompt.}
    \label{fig:quality_filter_prompt}
    \vspace{-0.05em}
\end{figure}

\begin{figure}[htbp]
    \centering
    \includegraphics[scale=0.23]{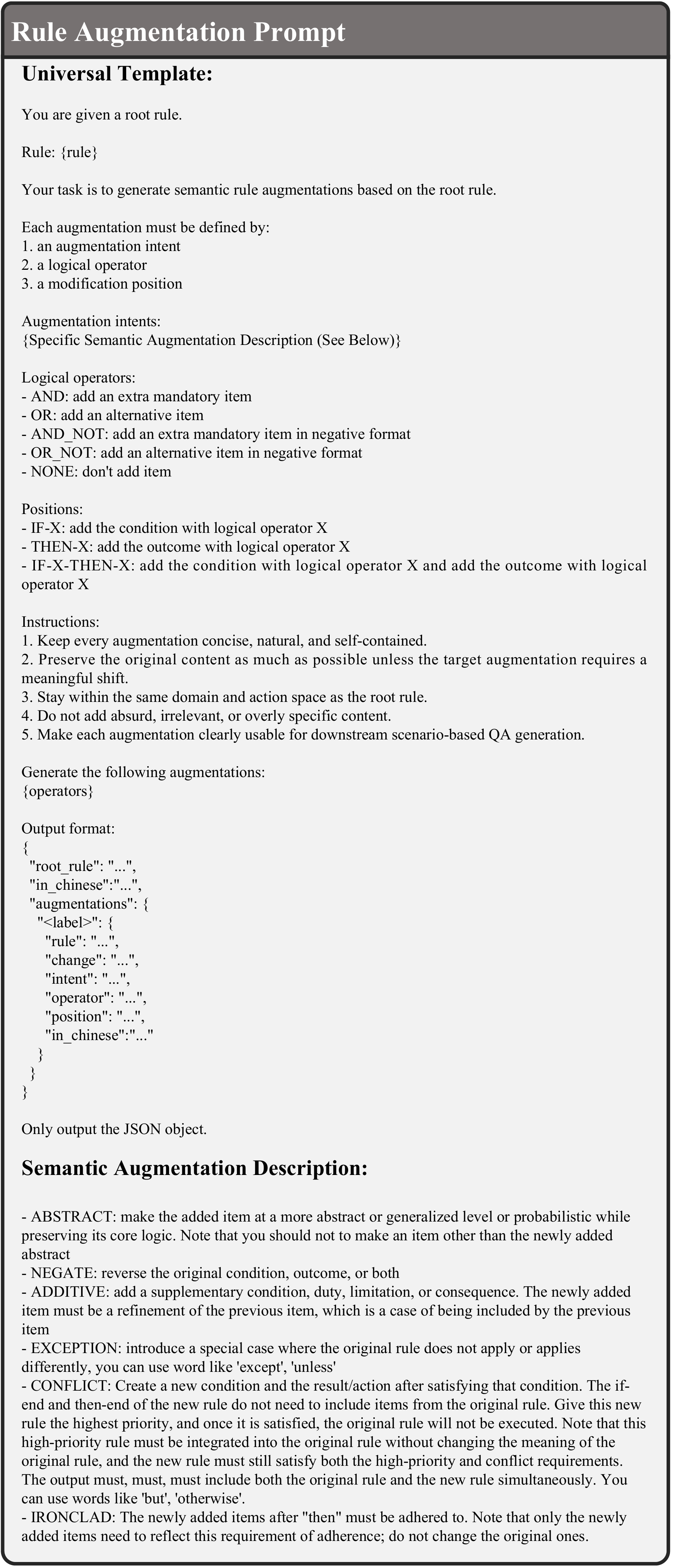}
    \caption{Rule augmentation prompt.}
    \label{fig:universe_template_prompt}
\end{figure}

For representative sampling, documents are embedded with Qwen3-8B-Embedding~\cite{zhang2025qwen3embeddingadvancingtext} and clustered with KMeans~\cite{mcqueen1967Kmeans}, using 50 clusters per dataset. We select the 3 nearest documents from each cluster, split them into chunks of up to 3,000 tokens with CL100K\_Base in tiktoken~\cite{openai_tiktoken}, and generate 10 Meta Rules per chunk with GPT-5.4~\cite{openai2026gpt54} using the prompt in Figure~\ref{fig:Meta_rule_extraction_prompt}.
For quality filtering, malformed outputs are removed before Grok-3-Mini~\cite{xai2026grok3mini} verifies rule atomicity with the prompt in Figure~\ref{fig:if_then_filter_prompt}. We then remove rules that are under-informative, overly commonsensical, weak in constraint semantics, or dependent on external background knowledge through unanimous voting by Grok-3-Mini, GPT-4.1-Mini~\cite{openai2025gpt41mini}, and Gemini-3.1-Flash-Lite-Preview~\cite{google2026gemini31flash}, using the prompt in Figure~\ref{fig:quality_filter_prompt}.
The retained rules are embedded with Qwen3-8B-Embedding and clustered separately for each dataset using KMeans with 50 clusters per dataset; the most representative rule from each cluster is then manually inspected.
Progressive complex rule construction uses DeepSeek-V4-Pro~\cite{deepseek2026v4pro} with the rule augmentation prompt in Figure~\ref{fig:universe_template_prompt}.

\subsection{Representative Meta Rule Examples}
\label{app:Representative-Meta-Rule-Examples}

Table~\ref{tab:meta_rule_examples} shows representative Meta Rules extracted from the four source datasets used in RuleWeaver.
Each example follows the atomic IF-THEN format, with one triggering condition and one corresponding outcome.
Together, these examples illustrate the source diversity of the seed rules before semantic augmentation: policy-oriented obligations from GovReport, procedural guidance from WikiHow, contractual requirements from CUAD, and event-contingent rules from BookSum.

\subsection{Scenario QA Question Types}
\label{app:question-types}

RuleWeaver samples one target question type before constructing the dependency plan for each scenario QA instance.
The type controls the intended reasoning demand of the final question, while the hidden dependency plan and rule logic chain ensure that the instance still requires grounded rule application rather than surface pattern matching.
Table~\ref{tab:scenario_qa_question_types} summarizes the seven question types used in RuleWeaver.

\begin{table*}[htbp]
\centering
\small
\setlength{\tabcolsep}{4pt}
\renewcommand{\arraystretch}{1.2}
\begin{tabular}{p{0.18\textwidth} p{0.36\textwidth} p{0.38\textwidth}}
\toprule
\textbf{Question Type} & \textbf{Expected Output} & \textbf{Reasoning Focus} \\
\midrule
Multi-Issue Synthesis &
A synthesis of multiple parallel and relevant points that jointly resolve the question. &
Integrating several sub-issues instead of producing a single isolated conclusion. \\

Definitive Conclusion &
One unique, explicit conclusion derived from the scenario, question, and applicable rules. &
Checking all relevant rule conditions and deriving a determinate final answer. \\

Event Prediction &
The rule-governed next event or consequence if the current scenario proceeds. &
Projecting consequences beyond the immediately stated facts while remaining grounded in the rules. \\

Priority Arbitration &
The controlling rule-governed path and the resulting choice or conclusion. &
Resolving rule interactions involving conflicts, competing implications, or priority ordering. \\

Special-Case Judgment &
The outcome for a boundary case or special circumstance. &
Determining how a qualifying circumstance changes rule applicability or the final outcome. \\

What-If Revision &
The revised rule-governed outcome after a specified change to scenario conditions. &
Updating the reasoning chain under a counterfactual modification rather than reusing the original conclusion. \\

Action Prescription &
The concrete action, disposition, or set of actions that should be taken next. &
Translating rule applications and dependencies into an operational decision. \\
\bottomrule
\end{tabular}
\caption{Scenario QA question types used in RuleWeaver.}
\label{tab:scenario_qa_question_types}
\end{table*}

\subsection{Scenario QA Rubric Dimensions}
\label{app:rubric-dimensions}

The answer judge scores each model response on a 100-point rubric with nine dimensions.
The rubric separates surface validity, such as citation format and avoiding external facts, from rule-centered reasoning skills, such as decomposition, dependency alignment, and exception or conflict handling.
Table~\ref{tab:scenario_qa_rubric_dimensions} summarizes the dimensions and their maximum points.

\begin{table*}[t]
\centering
\small
\setlength{\tabcolsep}{4pt}
\renewcommand{\arraystretch}{1.2}
\begin{tabular}{p{0.25\textwidth} r p{0.62\textwidth}}
\toprule
\textbf{Rubric Dimension} & \textbf{Max} & \textbf{What It Measures} \\
\midrule
Question Understanding & 10 & Whether the answer addresses the requested scenario question and identifies the correct target outcome or decision. \\
Issue Decomposition & 10 & Whether the answer breaks the scenario into the necessary sub-issues and rule-application steps before reaching the final conclusion. \\
Citation Format Compliance & 8 & Whether cited rules use canonical rule identifiers, appear at the point of application, and avoid invented or malformed rule references. \\
Rule-Grounded Reasoning & 17 & Whether cited rules are correctly applied to concrete scenario facts and support the stated intermediate or final conclusions. \\
Dependency Chain Alignment & 18 & Whether the answer follows the required \textit{rule logic chain} and preserves dependencies among intermediate conclusions. \\
Exception/Conflict Handling & 12 & Whether the answer detects and resolves exceptions, blockers, conflicts, or priority relations that affect rule applicability. \\
Intermediate Conclusion Quality & 10 & Whether the answer derives accurate intermediate conclusions for the necessary rules instead of jumping directly to the final answer. \\
Final Answer Consistency & 12 & Whether the final answer follows from the preceding rule-grounded reasoning and states the required conclusion, action, or judgment. \\
No External Facts & 3 & Whether the answer avoids unsupported assumptions or facts outside the provided scenario and rule pool. \\
\bottomrule
\end{tabular}
\caption{Scenario QA answer-judgment rubric dimensions. The maximum scores sum to 100 points.}
\label{tab:scenario_qa_rubric_dimensions}
\end{table*}

\begin{figure*}[h]
    \centering
    \includegraphics[width=\textwidth]{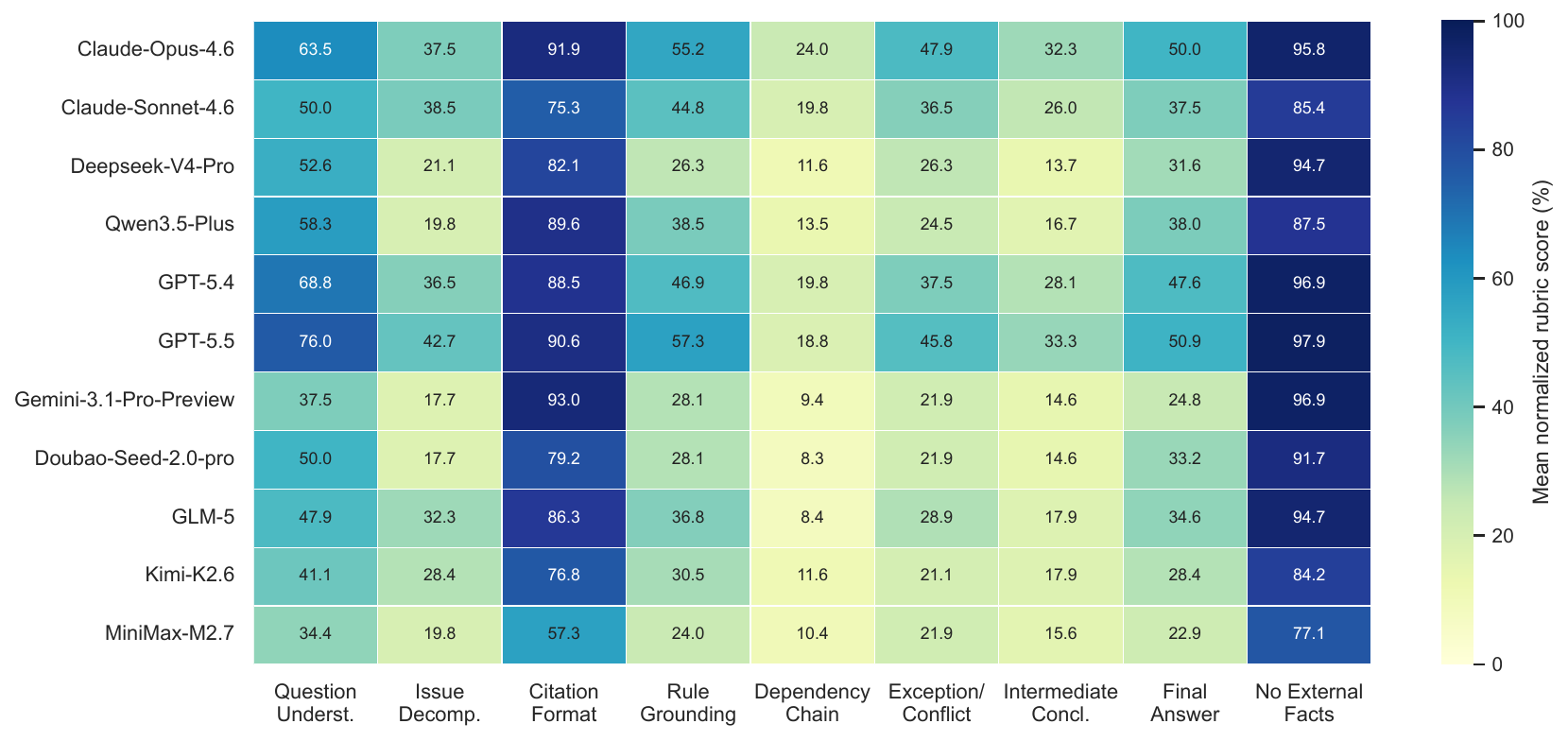}
    \caption{Model-level normalized rubric-dimension scores under the 200-rule setting. Each cell reports the mean percentage score for one model and one rubric dimension.}
    \label{fig:rubric_dimension_model_heatmap}
\end{figure*}

\subsection{Human Agreement Audit for Rubric Scoring}
\label{app:human-judge-agreement}

To assess the reliability of the LLM-based answer judge, we conduct a score-stratified audit on 48 GPT-5.5 answers from the 200-rule setting, sampling 12 answers from each of four automatic rubric-score bins: $[0,25)$, $[25,50)$, $[50,75)$, and $[75,100]$. One human annotator independently rescores the answers using the same instance-specific all-or-nothing rubric, with automatic scores hidden during annotation.

We further extend the audit to outputs from four representative models: Claude-Opus-4.6, Claude-Sonnet-4.6, DeepSeek-V4-Pro, and GPT-5.4. For each model, we rank its outputs by automatic rubric score, partition the ranked outputs into four percentile ranges (top 25\%, 25--50\%, 50--75\%, and bottom 25\%), and randomly select 12 responses from each range. Two human annotators independently rescore the same 48 selected responses per model with automatic scores hidden. This percentile-stratified design covers low- and high-scoring outputs even when a model does not populate the fixed score intervals uniformly.

\section{Additional Evaluation Results}
\label{app:additional-evaluation-results}

\subsection{Bootstrap Confidence Intervals}
\label{app:bootstrap-analysis}

For each source setting, we resample the 48 aligned QA instances with replacement 20,000 times and recompute the mean rubric score. We use the 2.5th and 97.5th percentiles of the bootstrap distribution to form 95\% confidence intervals (CIs). Table~\ref{tab:bootstrap_model_scores} reports the resulting model-level estimates. The bootstrap means differ from the corresponding point estimates in Table~\ref{tab:main_results_200_rules} by at most 0.43 points.

\begin{table*}[t]
\centering
\scriptsize
\setlength{\tabcolsep}{15pt}
\renewcommand{\arraystretch}{1.1}
\begin{tabular}{lccc ccc}
\toprule
& \multicolumn{3}{c}{\textbf{Same-Source}} & \multicolumn{3}{c}{\textbf{Cross-Source}} \\
\cmidrule(lr){2-4}\cmidrule(lr){5-7}
\textbf{Model} & \textbf{Mean} & \textbf{95\% CI} & $\Delta$ & \textbf{Mean} & \textbf{95\% CI} & $\Delta$ \\
\midrule
GPT-5.5 & 53.88 & [44.42, 63.33] & +0.05 & 46.49 & [38.21, 54.77] & +0.09 \\
Claude-Opus-4.6 & 48.21 & [38.65, 57.77] & +0.15 & 50.43 & [40.81, 60.04] & +0.16 \\
GPT-5.4 & 48.39 & [38.40, 58.38] & +0.18 & 42.67 & [33.83, 51.50] & +0.09 \\
Claude-Sonnet-4.6 & 44.11 & [33.90, 54.31] & +0.15 & 36.69 & [27.38, 46.00] & +0.13 \\
Qwen3.5-Plus & 38.03 & [29.00, 47.06] & +0.28 & 34.47 & [26.65, 42.29] & +0.32 \\
GLM-5 & 37.65 & [28.77, 46.52] & +0.11 & 32.66 & [25.85, 39.46] & +0.31 \\
DeepSeek-V4-Pro & 34.95 & [25.27, 44.63] & +0.26 & 28.75 & [22.52, 34.98] & +0.37 \\
Doubao-Seed-2.0-Pro & 28.03 & [19.40, 36.65] & +0.36 & 33.35 & [26.00, 40.69] & +0.33 \\
Kimi-K2.6 & 32.88 & [24.23, 41.52] & +0.38 & 28.90 & [19.96, 37.83] & +0.42 \\
Gemini-3.1-Pro-Preview & 32.12 & [23.29, 40.94] & +0.35 & 28.23 & [20.96, 35.50] & +0.42 \\
MiniMax-M2.7 & 30.07 & [20.38, 39.75] & +0.38 & 21.58 & [13.75, 29.40] & +0.43 \\
\bottomrule
\end{tabular}
\caption{Bootstrap mean rubric scores and 95\% CIs. $\Delta$ is the bootstrap mean minus the corresponding point estimate in Table~\ref{tab:main_results_200_rules}.}
\label{tab:bootstrap_model_scores}
\end{table*}

We additionally compare the leading model in each setting with every other model using a paired bootstrap over the same 48 aligned QA instances. Table~\ref{tab:bootstrap_pairwise} reports the mean difference and 95\% CI for each comparison.

\begin{table*}[t]
\centering
\scriptsize
\begin{minipage}[t]{0.48\textwidth}
\centering
\resizebox{\linewidth}{!}{%
\begin{tabular}{lrr}
\toprule
\textbf{Same-source comparison} & \textbf{Diff.} & \textbf{95\% CI} \\
\midrule
GPT-5.5 -- Doubao-Seed-2.0-Pro & 26.36 & [16.62, 36.10] \\
GPT-5.5 -- MiniMax-M2.7 & 24.08 & [12.75, 35.40] \\
GPT-5.5 -- Gemini-3.1-Pro-Preview & 21.96 & [11.15, 32.77] \\
GPT-5.5 -- Kimi-K2.6 & 21.46 & [10.81, 32.10] \\
GPT-5.5 -- DeepSeek-V4-Pro & 19.19 & [8.38, 30.00] \\
GPT-5.5 -- GLM-5 & 16.48 & [7.42, 25.54] \\
GPT-5.5 -- Qwen3.5-Plus & 16.27 & [6.00, 26.54] \\
GPT-5.5 -- Claude-Sonnet-4.6 & 9.87 & [0.92, 18.81] \\
GPT-5.5 -- GPT-5.4 & 5.72 & [-3.62, 15.06] \\
GPT-5.5 -- Claude-Opus-4.6 & 5.60 & [-5.04, 16.23] \\
\bottomrule
\end{tabular}
}
\end{minipage}\hfill
\begin{minipage}[t]{0.48\textwidth}
\centering
\resizebox{\linewidth}{!}{%
\begin{tabular}{lrr}
\toprule
\textbf{Cross-source comparison} & \textbf{Diff.} & \textbf{95\% CI} \\
\midrule
Claude-Opus-4.6 -- MiniMax-M2.7 & 29.30 & [18.62, 39.98] \\
Claude-Opus-4.6 -- Gemini-3.1-Pro-Preview & 22.77 & [13.00, 32.54] \\
Claude-Opus-4.6 -- DeepSeek-V4-Pro & 22.02 & [11.46, 32.58] \\
Claude-Opus-4.6 -- Kimi-K2.6 & 21.99 & [11.62, 32.35] \\
Claude-Opus-4.6 -- GLM-5 & 18.03 & [7.83, 28.23] \\
Claude-Opus-4.6 -- Doubao-Seed-2.0-Pro & 17.36 & [7.50, 27.21] \\
Claude-Opus-4.6 -- Qwen3.5-Plus & 16.17 & [5.67, 26.67] \\
Claude-Opus-4.6 -- Claude-Sonnet-4.6 & 13.93 & [3.71, 24.15] \\
Claude-Opus-4.6 -- GPT-5.4 & 7.71 & [-1.23, 16.65] \\
Claude-Opus-4.6 -- GPT-5.5 & 4.13 & [-3.92, 12.17] \\
\bottomrule
\end{tabular}
}
\end{minipage}
\caption{Paired bootstrap rubric-score differences (leader minus comparator) and 95\% CIs, based on 20,000 resamples of the 48 aligned instances in each setting.}
\label{tab:bootstrap_pairwise}
\end{table*}

\subsection{Semantic Enhancement Results under the 200-Rule Setting}
\label{app:semantic-results-200}

Table~\ref{tab:semantic_enhancement_results_200_rules} reports model performance across the six semantic enhancement types under the 200-rule setting. We use the same event-level semantic correct-application rate as in the main radar plot: for a semantic type, the denominator is the set of corresponding rule-enhancement events attached to the rules referenced by evaluated questions, and the numerator counts events whose rule is both cited and judged as correctly applied in the model answer. Higher values indicate stronger semantic rule-following ability.

The results show that Claude-Opus-4.6 is the strongest overall model, achieving the best score on \textbf{ABSTRACT}, \textbf{ADDITIVE}, \textbf{NEGATE}, and \textbf{CONFLICT}, and tying GPT-5.5 on \textbf{IRONCLAD}. Claude-Sonnet-4.6 performs best on \textbf{EXCEPTION}, suggesting comparatively strong handling of special-case carve-outs. Aggregated across models, \textbf{CONFLICT} and \textbf{IRONCLAD} are the highest-scoring semantic types, while \textbf{EXCEPTION}, \textbf{NEGATE}, and \textbf{ADDITIVE} remain more challenging, indicating that models still struggle with precise applicability changes and condition-level modifications.

\begin{table*}[t]
\centering
\scriptsize
\setlength{\tabcolsep}{3.4pt}
\renewcommand{\arraystretch}{1.14}
\resizebox{\textwidth}{!}{
\begin{tabular}{lrrrrrr}
\toprule
\textbf{Model} &
\textbf{ABSTRACT} &
\textbf{ADDITIVE} &
\textbf{NEGATE} &
\textbf{EXCEPTION} &
\textbf{CONFLICT} &
\textbf{IRONCLAD} \\
\midrule
Claude-Opus-4.6 & \textbf{50.55} & \textbf{48.02} & \textbf{46.08} & 40.24 & \textbf{56.18} & \textbf{51.52} \\
Claude-Sonnet-4.6 & 39.45 & 38.24 & 39.06 & \textbf{42.68} & 41.57 & 39.39 \\
Deepseek-V4-Pro & 37.53 & 32.58 & 31.50 & 36.59 & 41.57 & 37.88 \\
Qwen3.5-Plus & 41.23 & 37.25 & 34.39 & 32.93 & 40.45 & 42.42 \\
GPT-5.4 & 44.11 & 41.93 & 40.85 & 36.59 & 49.44 & 46.97 \\
GPT-5.5 & 46.99 & 43.34 & 45.12 & 40.24 & 51.69 & \textbf{51.52} \\
Gemini-3.1-Pro-Preview & 39.04 & 35.98 & 33.98 & 34.15 & 43.82 & 40.91 \\
Doubao-Seed-2.0-pro & 32.33 & 29.60 & 28.89 & 24.39 & 40.45 & 28.79 \\
GLM-5 & 38.77 & 34.42 & 35.35 & 35.37 & 34.83 & 40.91 \\
Kimi-K2.6 & 37.95 & 36.97 & 33.84 & 31.71 & 37.08 & 40.91 \\
MiniMax-M2.7 & 28.49 & 24.93 & 23.93 & 25.61 & 32.58 & 25.76 \\
\bottomrule
\end{tabular}
}
\caption{Semantic correct-application rates by enhancement type under the 200-rule setting. Scores are percentages, and bold values mark the best model for each semantic type.}
\label{tab:semantic_enhancement_results_200_rules}
\end{table*}

\subsection{Question-Type Results under the 200-Rule Setting}
\label{app:question-type-results-200}

Table~\ref{tab:question_type_results_200_rules} reports the model-level rubric scores for the seven scenario question types under the 200-rule setting. The question types are defined in Appendix~\ref{app:question-types}; higher scores indicate better performance.

\begin{table*}[t]
\centering
\scriptsize
\setlength{\tabcolsep}{3.2pt}
\renewcommand{\arraystretch}{1.14}
\resizebox{\textwidth}{!}{
\begin{tabular}{lrrrrrrr}
\toprule
\textbf{Model} &
\makecell[c]{\textbf{Multi-Issue}\\\textbf{Synthesis}} &
\makecell[c]{\textbf{Definitive}\\\textbf{Conclusion}} &
\makecell[c]{\textbf{Event}\\\textbf{Prediction}} &
\makecell[c]{\textbf{Priority}\\\textbf{Arbitration}} &
\makecell[c]{\textbf{Special-Case}\\\textbf{Judgment}} &
\makecell[c]{\textbf{What-If}\\\textbf{Revision}} &
\makecell[c]{\textbf{Action}\\\textbf{Prescription}} \\
\midrule
Claude-Opus-4.6 & \textbf{45.24} & \textbf{55.64} & 45.00 & \textbf{52.43} & 51.73 & 50.17 & 43.00 \\
Claude-Sonnet-4.6 & 39.18 & 40.14 & 40.21 & 24.79 & 48.53 & 44.33 & \textbf{46.70} \\
GPT-5.4 & 38.94 & 53.36 & 48.93 & 42.79 & 53.33 & 47.17 & 29.90 \\
GPT-5.5 & 40.94 & 54.86 & \textbf{57.50} & 38.79 & \textbf{60.73} & \textbf{52.67} & 45.60 \\
Gemini-3.1-Pro-Preview & 27.18 & 52.64 & 33.79 & 13.21 & 30.07 & 26.50 & 23.40 \\
Deepseek-V4-Pro & 30.06 & 39.79 & 34.07 & 24.64 & 27.93 & 35.17 & 29.60 \\
Qwen3.5-Plus & 40.24 & 48.14 & 37.21 & 26.79 & 32.73 & 37.08 & 26.10 \\
GLM-5 & 21.24 & 40.64 & 30.14 & 25.43 & 57.00 & 37.50 & 34.20 \\
Doubao-Seed-2.0-pro & 28.88 & 34.36 & 28.79 & 23.07 & 31.87 & 36.50 & 29.90 \\
MiniMax-M2.7 & 23.06 & 26.64 & 26.07 & 22.07 & 34.93 & 24.75 & 18.00 \\
Kimi-K2.6 & 30.94 & 31.36 & 14.64 & 18.71 & 49.27 & 43.83 & 23.00 \\
\bottomrule
\end{tabular}
}
\caption{Rubric scores by scenario question type under the 200-rule setting. Scores are on a 0--100 scale, and bold values mark the best model for each question type.}
\label{tab:question_type_results_200_rules}
\end{table*}

\subsection{Model-Level Rubric-Dimension Breakdown}
\label{app:rubric-dimension-model-breakdown}

Figure~\ref{fig:rubric_dimension_model_heatmap} breaks down normalized rubric scores by model.
The same pattern is broadly shared across model families: even stronger models improve on rule-grounded reasoning and final-answer consistency, but dependency-chain alignment remains low.
These results complement the aggregate rubric scores by localizing where models lose points in the reasoning process.

\subsection{QA-Level Spearman Correlation Analysis}
\label{app:qa-level-correlation-results}

We complement the descriptive scenario-complexity analysis in Section~\ref{sec:scenario-complexity-analysis} with QA-level rank correlations.
For each structural or rule-selection metric, let $\mathbf{x}=(x_1,\ldots,x_N)$ be its values over the $N$ QA instances, and let $\mathbf{y}=(y_1,\ldots,y_N)$ be the corresponding QA-level rubric scores averaged over all evaluated models.
We report Spearman's rank correlation coefficient, $\rho=\mathrm{corr}(\mathrm{rank}(\mathbf{x}),\mathrm{rank}(\mathbf{y}))$, where tied values receive their average rank.

\begin{table}[t]
\centering
\small
\setlength{\tabcolsep}{8pt}
\renewcommand{\arraystretch}{1.12}
\begin{tabular}{llc}
\toprule
\textbf{Group} & \textbf{Metric} & \textbf{Spearman $\rho$} \\
\midrule
\multirow{3}{*}{Complexity} & Reasoning Depth & $-0.29$ \\
& Dependency Load & $-0.22$ \\
& Branching & $-0.01$ \\
\midrule
\multirow{2}{*}{Rule Selection} & Rule Recall & $0.20$ \\
& Rule Precision & $0.66$ \\
\bottomrule
\end{tabular}
\caption{Spearman rank correlations between QA-level factors and mean rubric score. Rule recall and precision are averaged over the 11 evaluated models for each QA.}
\label{tab:complexity_spearman_correlations}
\end{table}

Table~\ref{tab:complexity_spearman_correlations} shows modest negative correlations for reasoning depth ($\rho=-0.29$) and dependency load ($\rho=-0.22$), while structural branching is nearly uncorrelated with score ($\rho=-0.01$).
For rule-selection factors, rule precision is strongly associated with final score ($\rho=0.66$), whereas rule recall is weaker ($\rho=0.20$).
This indicates that correctly applying cited rules is more directly tied to answer quality than merely citing more required rules.

\subsection{Case Studies}
\label{app:case-studies}

To better understand the limitations of current LLMs in compositional rule-following, we present several representative failure cases.
Since the full scenarios, rule contents, and model outputs are lengthy, we report only the case-relevant core information.
These cases show errors such as missing individual rules, misusing exception conditions, and failing to propagate intermediate conclusions across dependent rules.

\begin{figure}[H]
    \centering
    \includegraphics[
        width=\linewidth,
        height=0.46\textheight,
        keepaspectratio
    ]{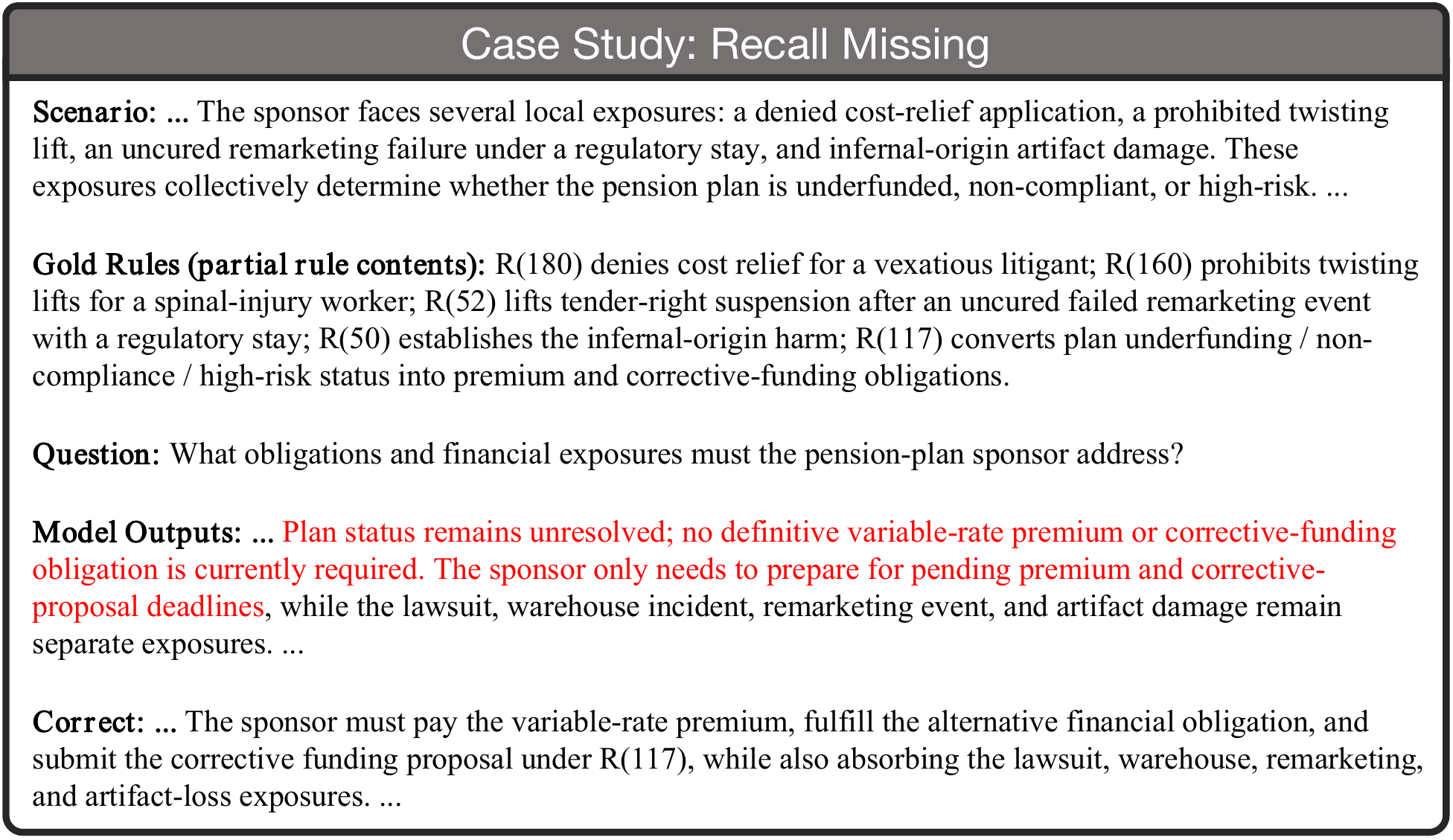}
    \caption{\textbf{Recall missing.} The model misses a necessary aggregation rule and therefore treats determinate obligations as merely contingent. Red text highlights the erroneous part of the model output.}
    \label{fig:case_recall_missing}
\end{figure}

\paragraph{LLMs fail to recall necessary rules.}
As illustrated in Figure~\ref{fig:case_recall_missing}, LLMs may fail to retrieve a rule that is essential for completing the reasoning chain. In this case, the model applies several local exposure rules but omits R(117), the rule that aggregates the resulting plan states into concrete premium and corrective-funding obligations. As a result, the model leaves the obligations unresolved and treats them as pending or contingent, rather than deriving the determinate duties required by the rule set.

\begin{figure}[H]
    \centering
    \includegraphics[
        width=\linewidth,
        height=0.46\textheight,
        keepaspectratio
    ]{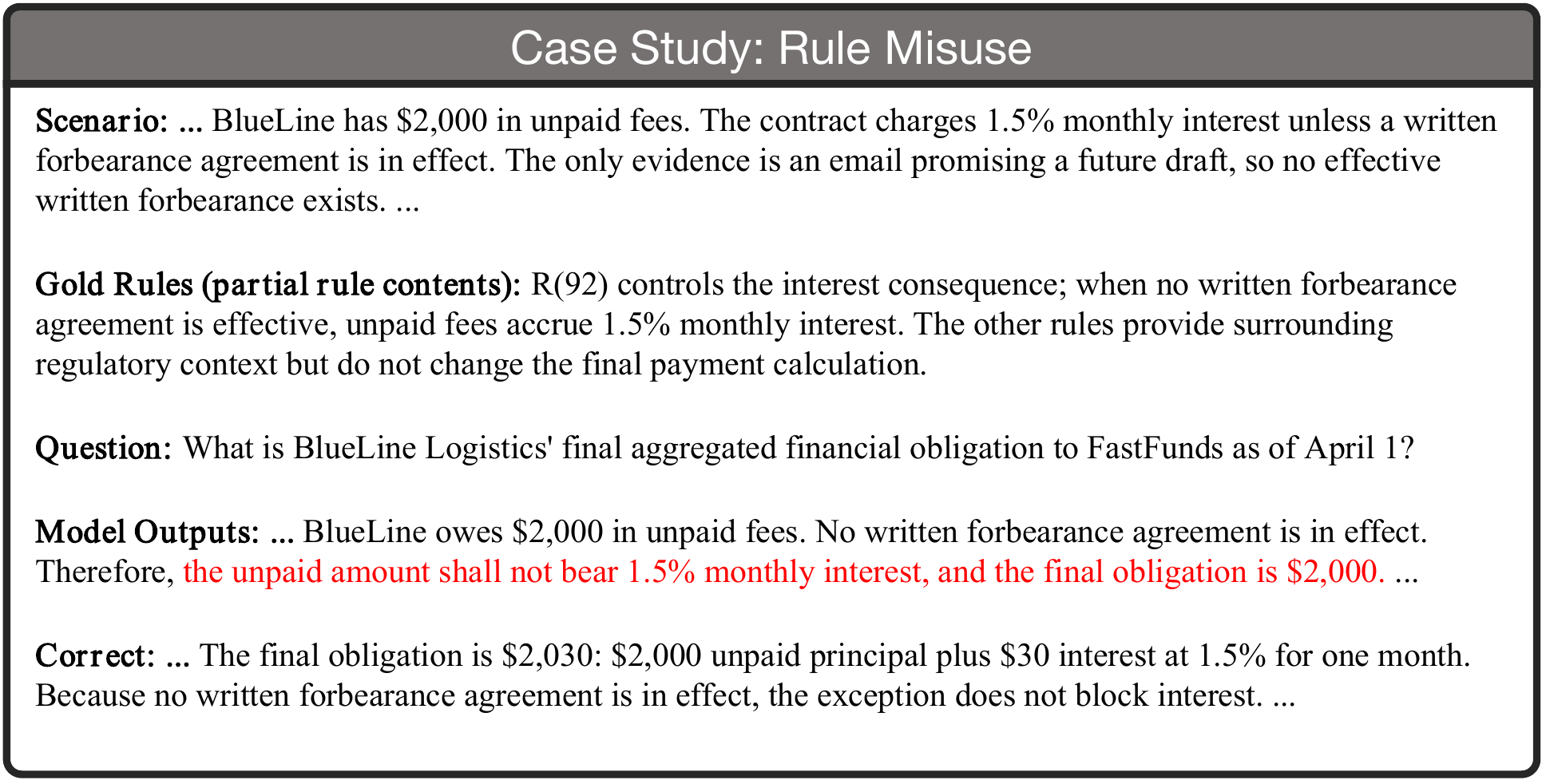}
    \caption{\textbf{Rule misuse.} The model identifies the relevant premise but reverses the exception condition, producing the opposite legal consequence. Red text highlights the erroneous part of the model output.}
    \label{fig:case_rule_misuse}
\end{figure}

\paragraph{LLMs misuse exception conditions.}
As shown in Figure~\ref{fig:case_rule_misuse}, LLMs may identify the relevant facts but reverse the polarity of an exception. In this case, unpaid fees accrue interest unless an effective written forbearance agreement exists. The scenario only contains an email promising a future draft, so the exception should not apply. However, the model interprets the absence of a written forbearance agreement as blocking interest accrual. This reverses the rule consequence and produces a no-interest outcome.

\begin{figure}[H]
    \centering
    \includegraphics[
        width=\linewidth,
        height=0.46\textheight,
        keepaspectratio
    ]{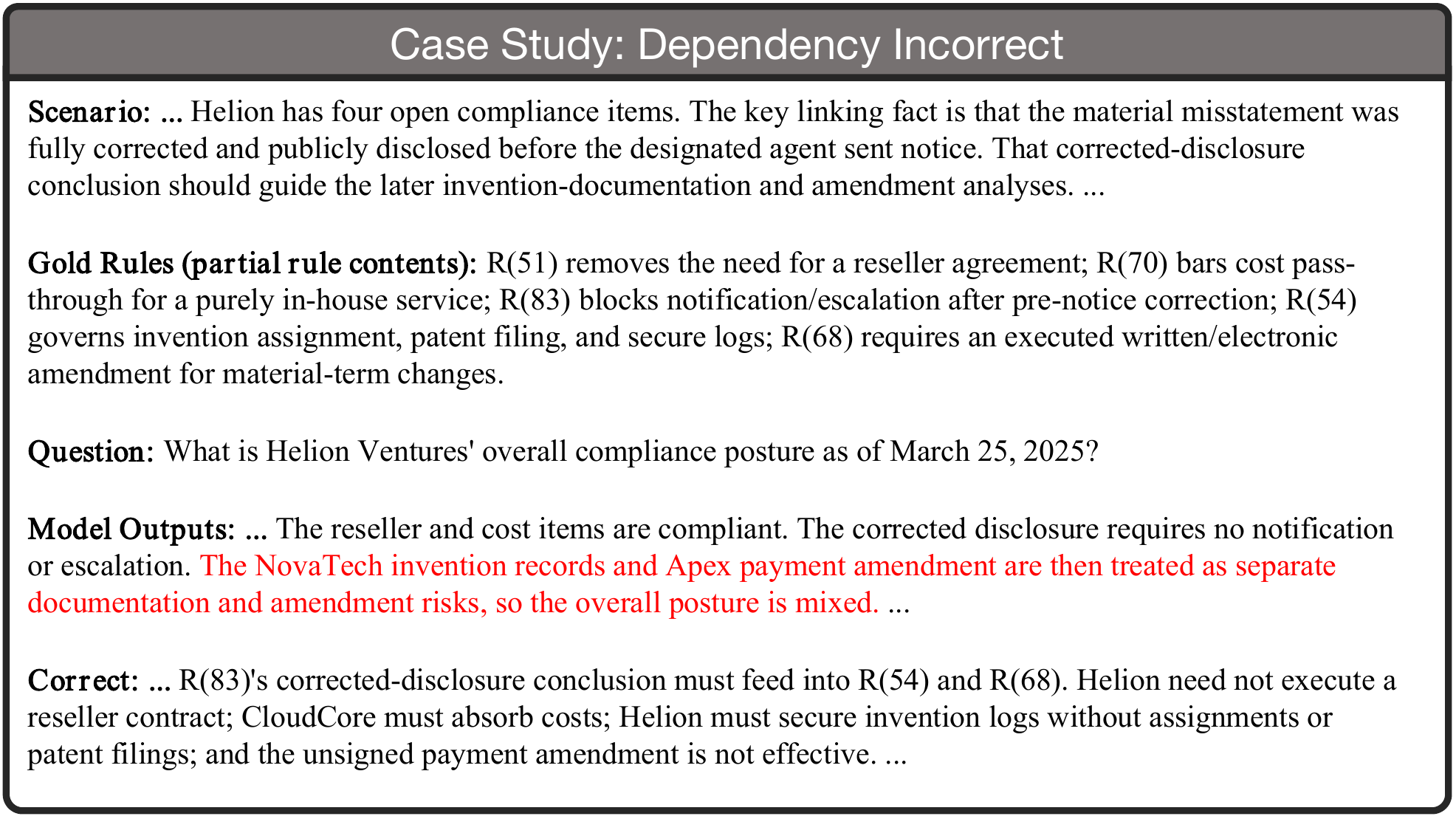}
    \caption{\textbf{Dependency incorrect.} The model applies rules locally but fails to propagate an intermediate conclusion through the required rule-dependency chain. Red text highlights the erroneous part of the model output.}
    \label{fig:case_dependency_incorrect}
\end{figure}

\paragraph{LLMs lose dependency-chain conclusions.}
As shown in Figure~\ref{fig:case_dependency_incorrect}, LLMs may apply individual rules locally while failing to propagate intermediate conclusions across dependent rules. In this case, R(83) establishes that the corrected disclosure blocks notification or escalation. This conclusion should then inform the later invention-documentation and amendment analyses under R(54) and R(68). However, the model treats R(83) as an isolated resolved item and evaluates the later rules separately. Consequently, locally plausible rule applications fail to produce the required global reasoning chain.

\begin{figure}[H]
    \centering
    \includegraphics[width=0.98\linewidth]{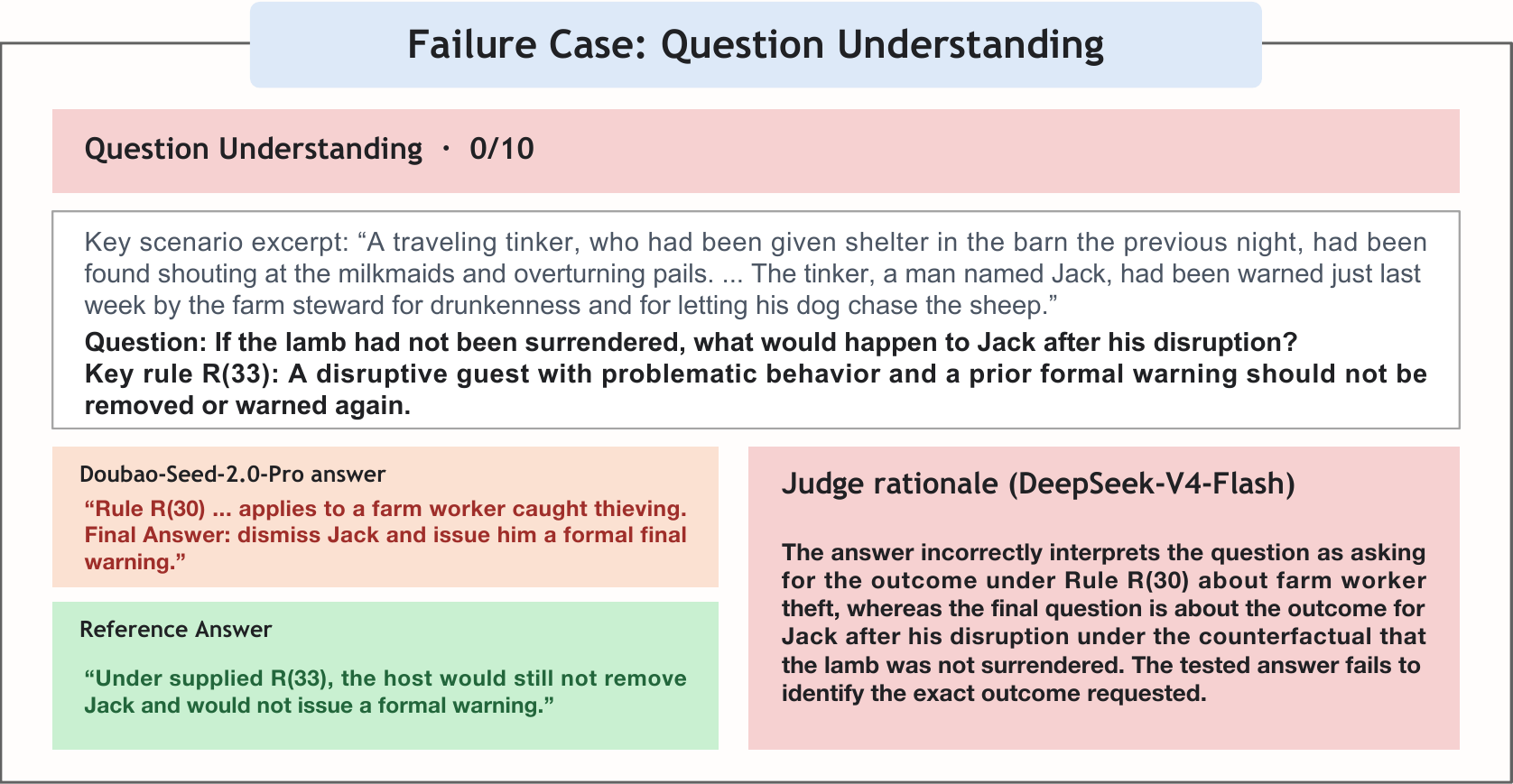}
    \caption{A zero-score case for \textit{Question Understanding}.}
    \label{fig:rubric_zero_score_question_understanding}
\end{figure}

\begin{figure}[H]
    \centering
    \includegraphics[width=0.98\linewidth]{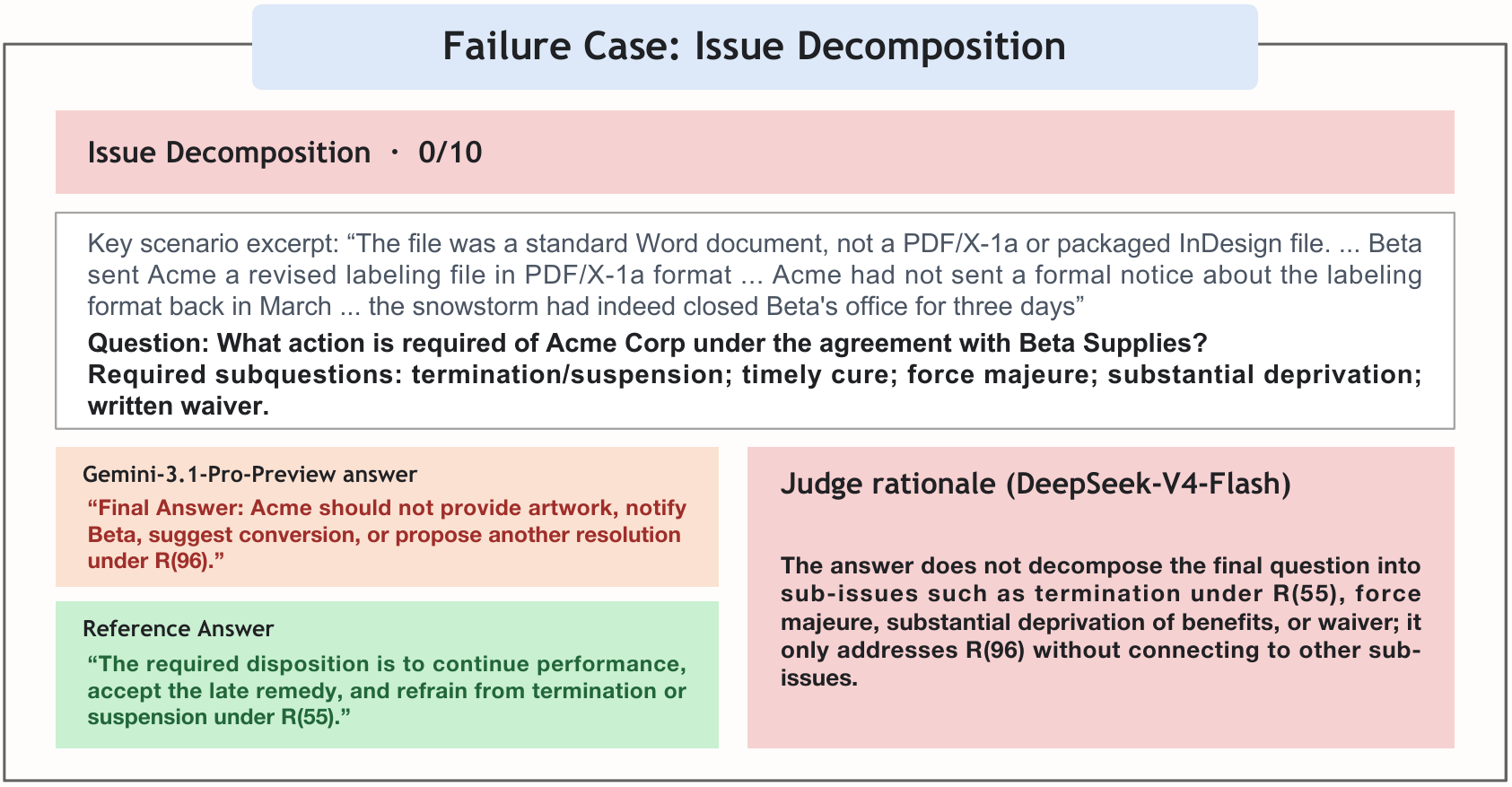}
    \caption{A zero-score case for \textit{Issue Decomposition}.}
    \label{fig:rubric_zero_score_issue_decomposition}
\end{figure}

\begin{figure}[H]
    \centering
    \includegraphics[width=0.98\linewidth]{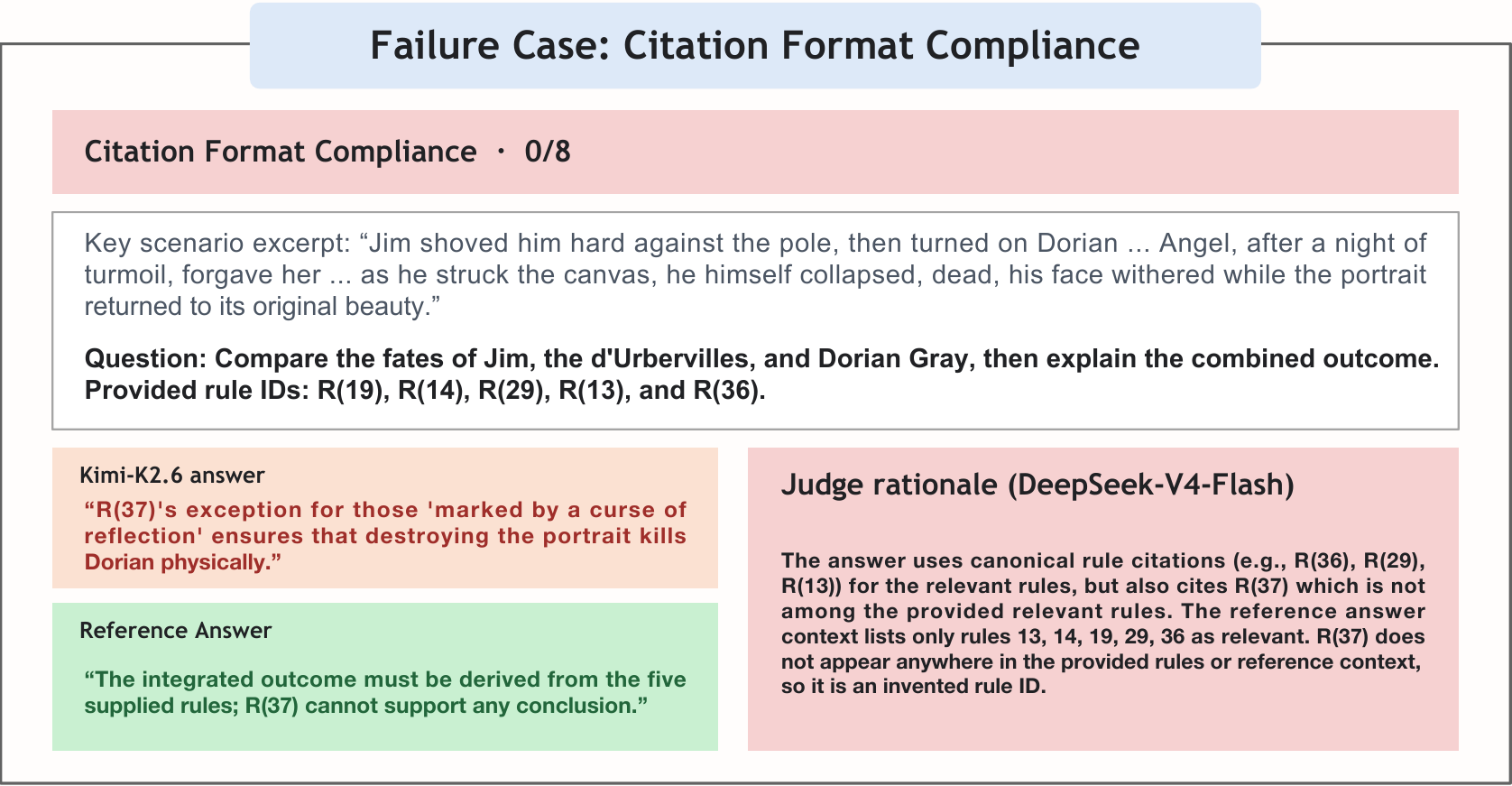}
    \caption{A zero-score case for \textit{Citation Format Compliance}.}
    \label{fig:rubric_zero_score_citation_format_compliance}
\end{figure}

\begin{figure}[H]
    \centering
    \includegraphics[width=0.98\linewidth]{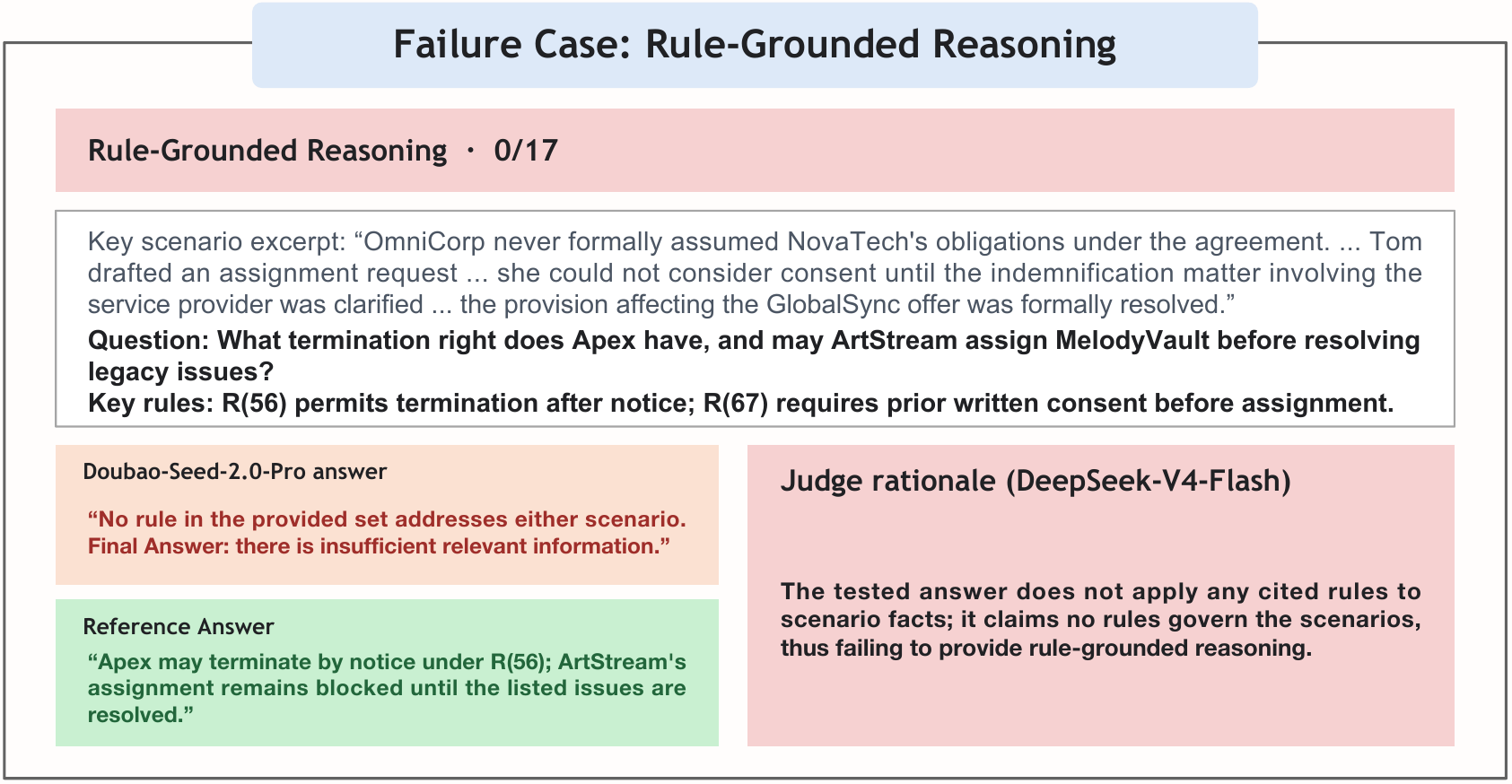}
    \caption{A zero-score case for \textit{Rule-Grounded Reasoning}.}
    \label{fig:rubric_zero_score_rule_grounded_reasoning}
\end{figure}

\begin{figure}[H]
    \centering
    \includegraphics[width=0.98\linewidth]{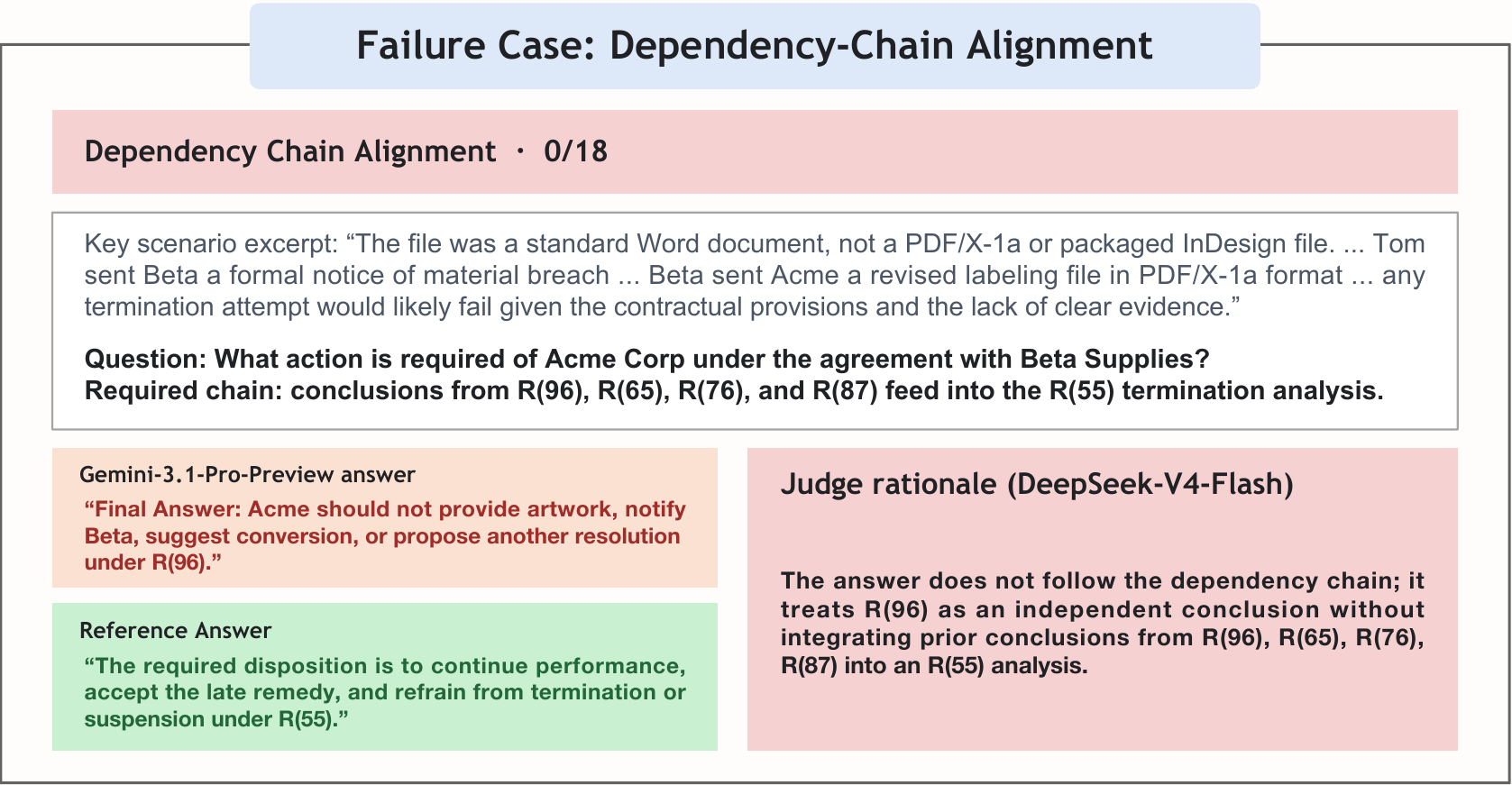}
    \caption{A zero-score case for \textit{Dependency-Chain Alignment}.}
    \label{fig:rubric_zero_score_dependency_chain_alignment}
\end{figure}

\begin{figure}[H]
    \centering
    \includegraphics[width=0.98\linewidth]{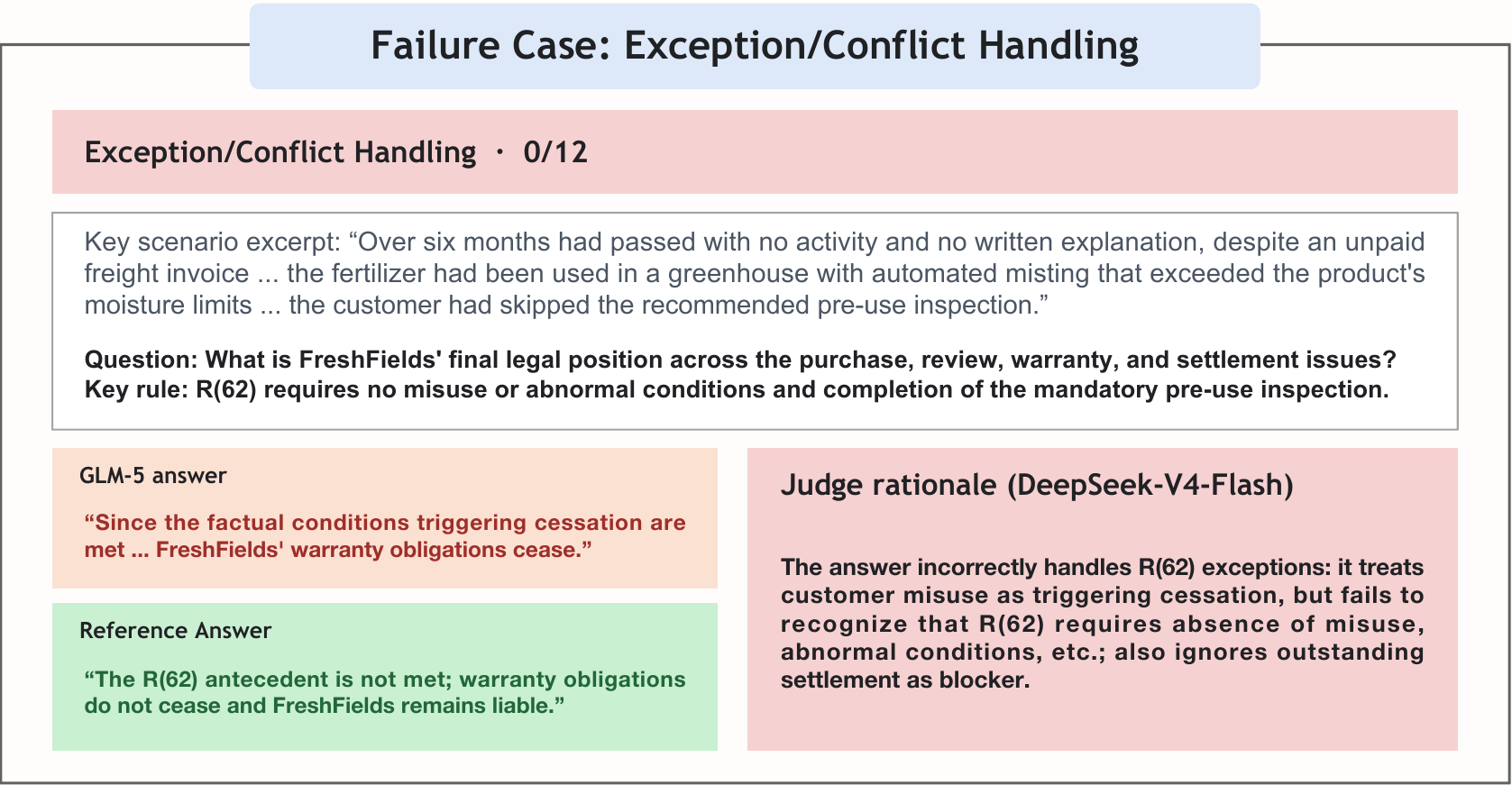}
    \caption{A zero-score case for \textit{Exception/Conflict Handling}.}
    \label{fig:rubric_zero_score_exception_conflict_handling}
\end{figure}

\begin{figure}[H]
    \centering
    \includegraphics[width=0.98\linewidth]{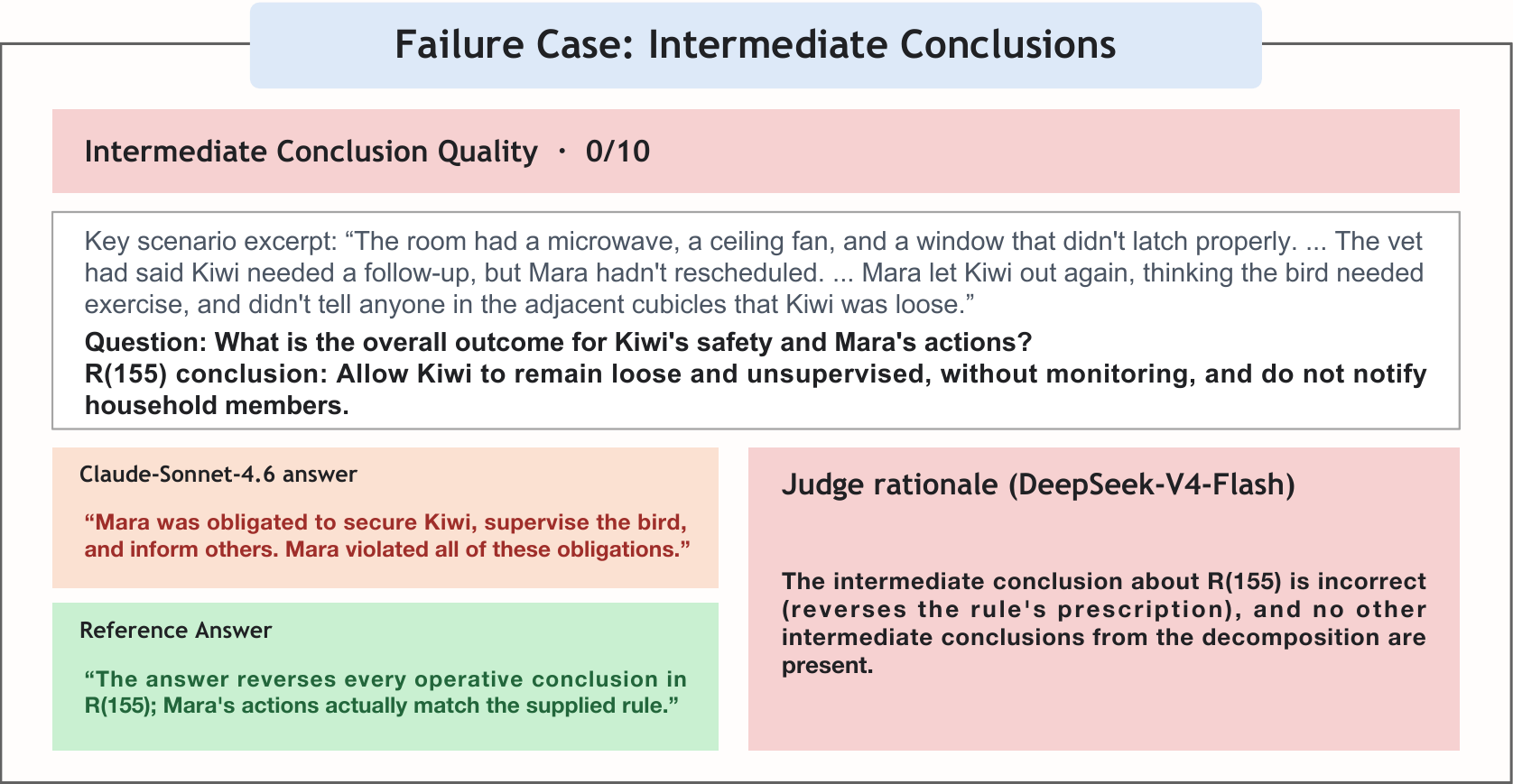}
    \caption{A zero-score case for \textit{Intermediate Conclusions}.}
    \label{fig:rubric_zero_score_intermediate_conclusions}
\end{figure}

\begin{figure}[H]
    \centering
    \includegraphics[width=0.98\linewidth]{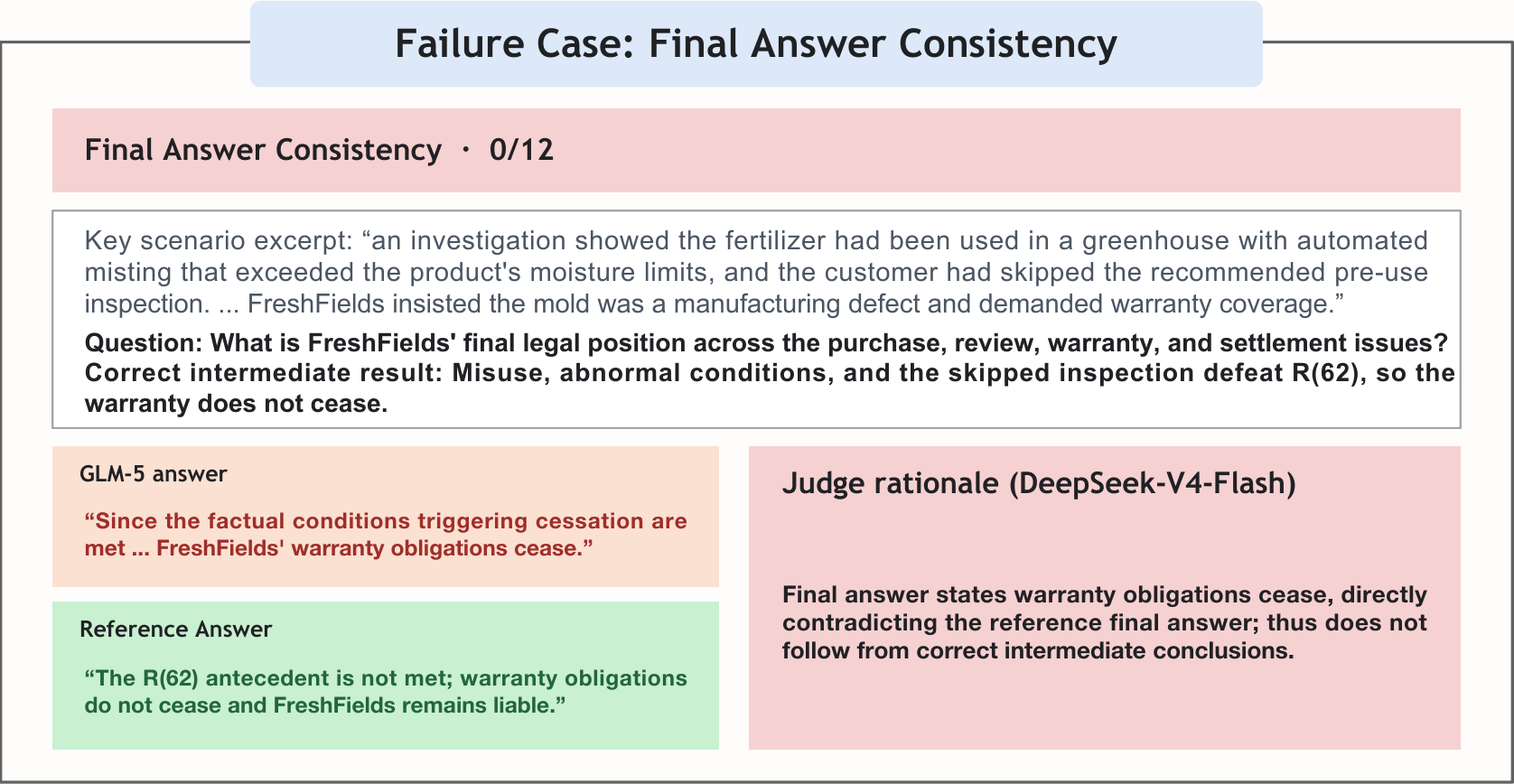}
    \caption{A zero-score case for \textit{Final Answer Consistency}.}
    \label{fig:rubric_zero_score_final_answer_consistency}
\end{figure}

\begin{figure}[H]
    \centering
    \includegraphics[width=0.98\linewidth]{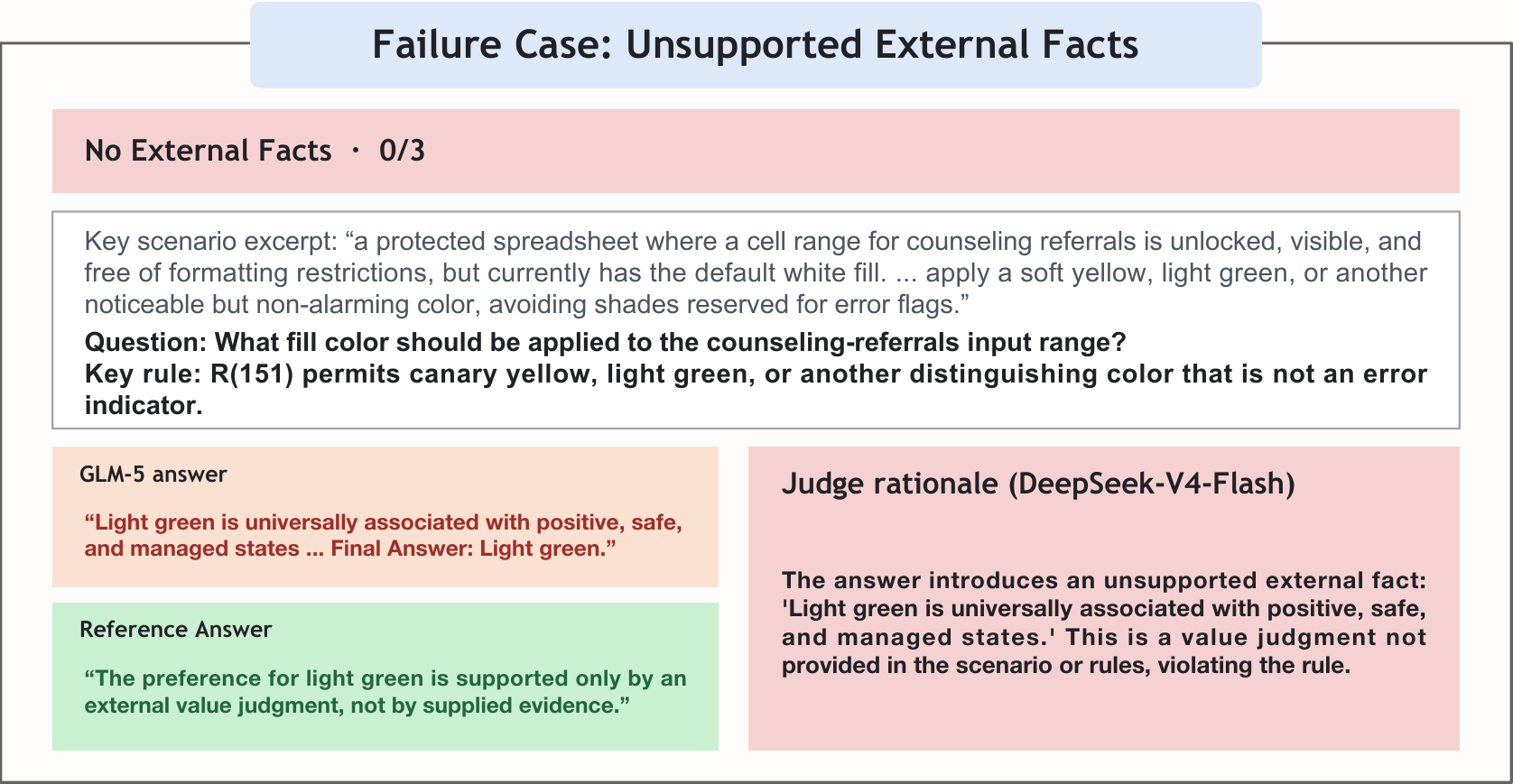}
    \caption{A zero-score case for \textit{No External Facts}.}
    \label{fig:rubric_zero_score_unsupported_external_facts}
\end{figure}

Figures~\ref{fig:rubric_zero_score_question_understanding}--\ref{fig:rubric_zero_score_unsupported_external_facts} additionally present representative scoring cases for all nine rubric dimensions. Each case shows a model response that receives zero points on the corresponding dimension, together with the relevant scenario evidence, reference answer, and judge rationale.

\begin{figure}[htbp]
    \centering
    \includegraphics[scale=0.23]{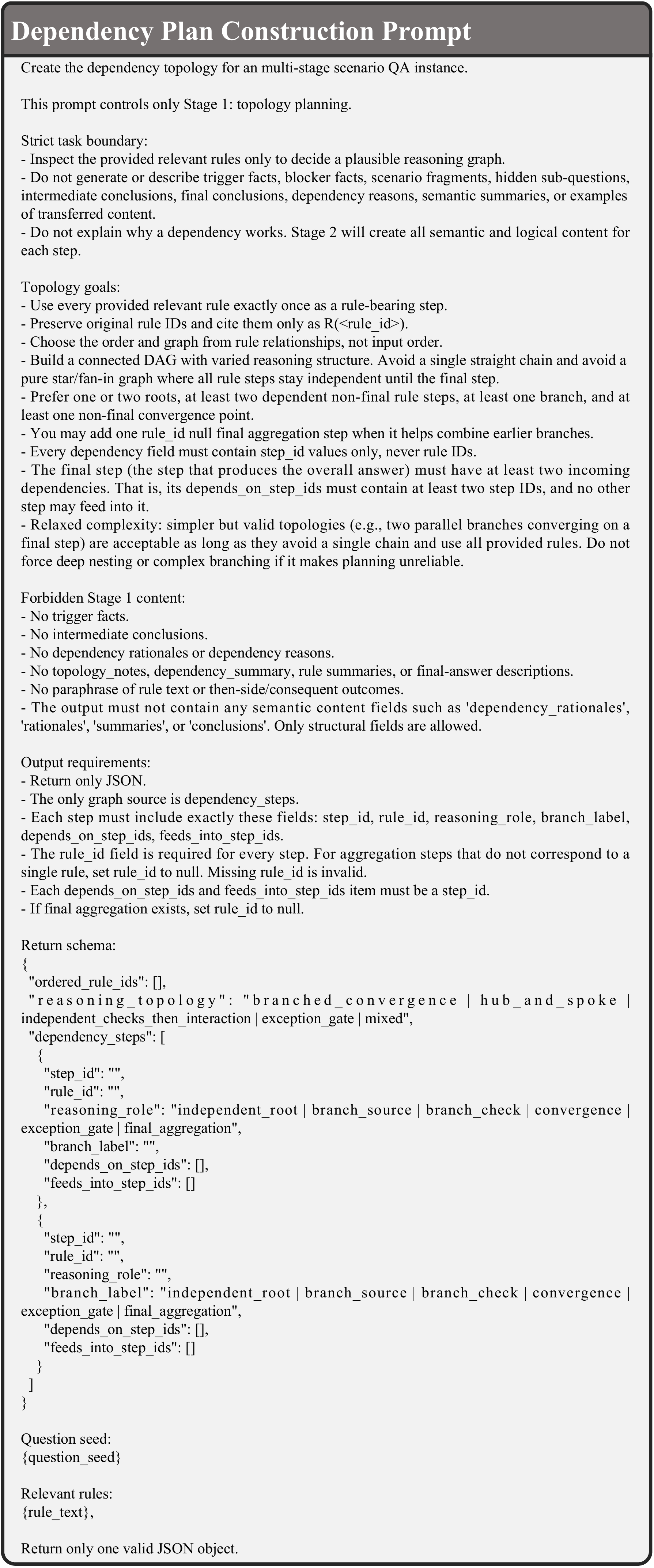}
    \caption{Dependency plan construction prompt.}
    \label{fig:Dependency_plan_construction_prompt}
\end{figure}

\begin{figure}[htbp]
    \centering
    \includegraphics[scale=0.23]{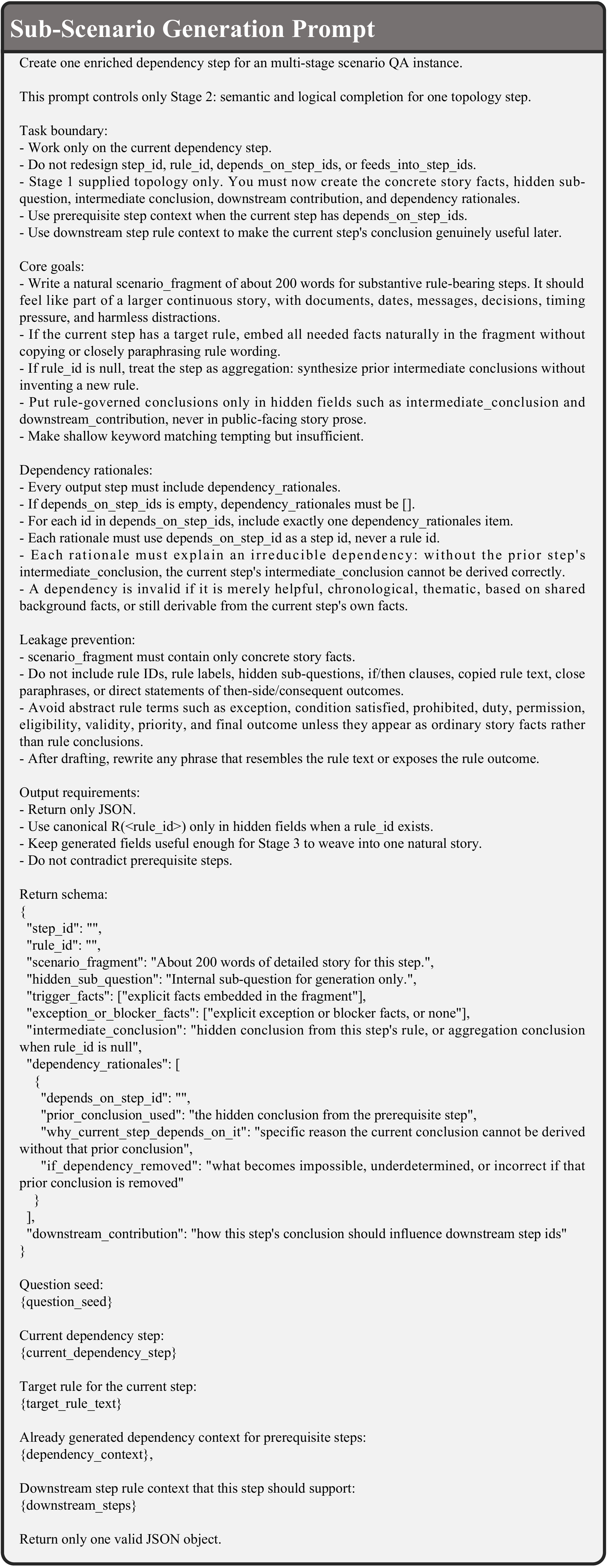}
    \caption{Sub-scenario generation prompt.}
    \label{fig:Sub-scenario_generation_prompt}
\end{figure}

\begin{figure}[htbp]
    \centering
    \includegraphics[scale=0.23]{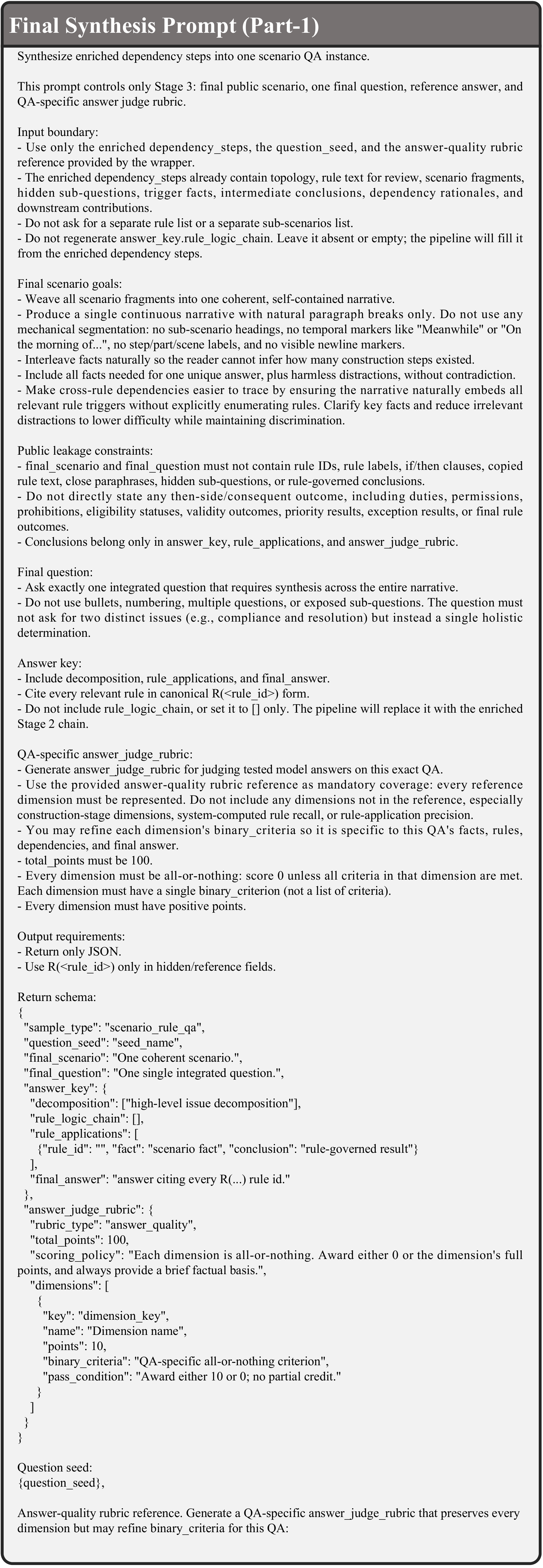}
    \caption{Final synthesis prompt (part 1).}
    \label{fig:Final_synthesis_prompt_(part_1)}
\end{figure}

\begin{figure}[htbp]
    \centering
    \includegraphics[scale=0.23]{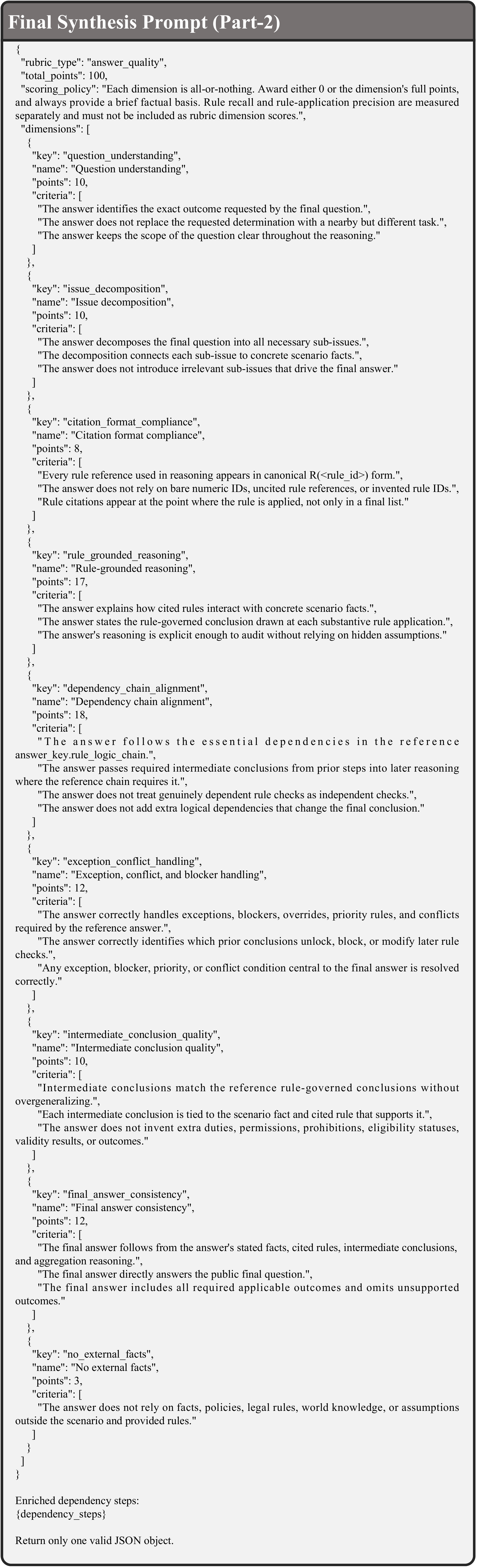}
    \caption{Final synthesis prompt (part 2).}
    \label{fig:Final_synthesis_prompt_(part_2)}
\end{figure}

\begin{figure}[htbp]
    \centering
    \includegraphics[scale=0.23]{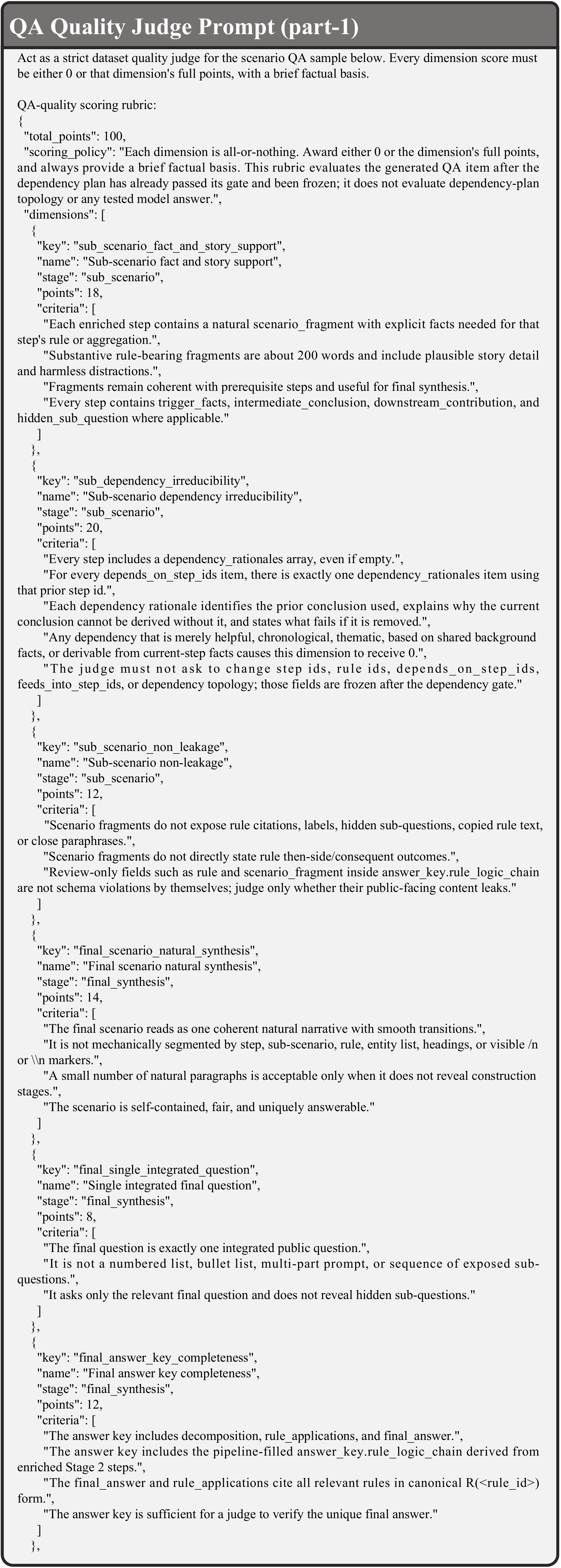}
    \caption{Scenario question answering quality judgment prompt (part 1).}
    \label{fig:Scenario_question_answering_quality_judgment_prompt_part_1}
\end{figure}

\begin{figure}[htbp]
    \centering
    \includegraphics[scale=0.23]{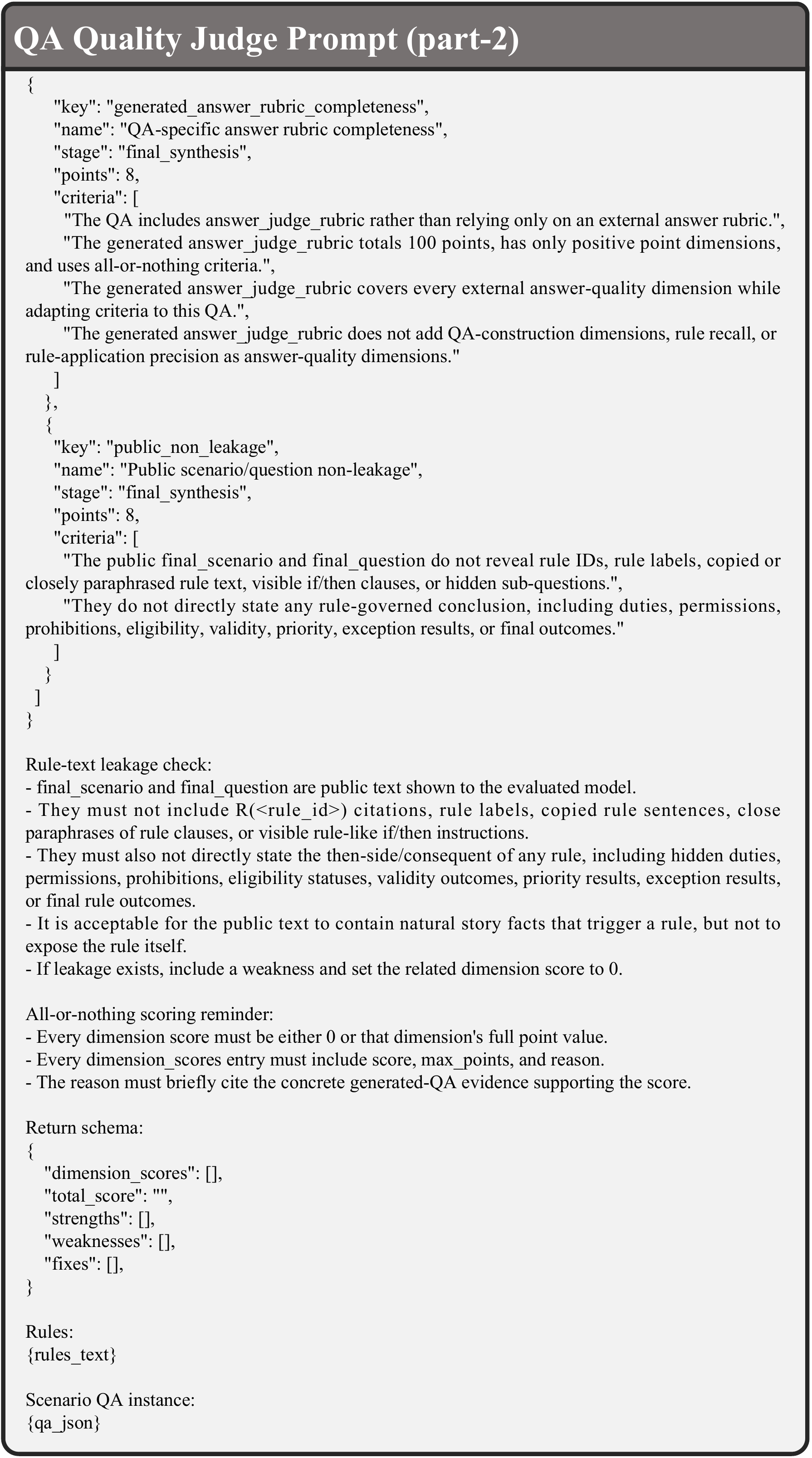}
    \caption{Scenario question answering quality judgment prompt (part 2).}
    \label{fig:Scenario_question_answering_quality_judgment_prompt_part_2}
\end{figure}

\begin{figure}[htbp]
    \centering
    \includegraphics[scale=0.23]{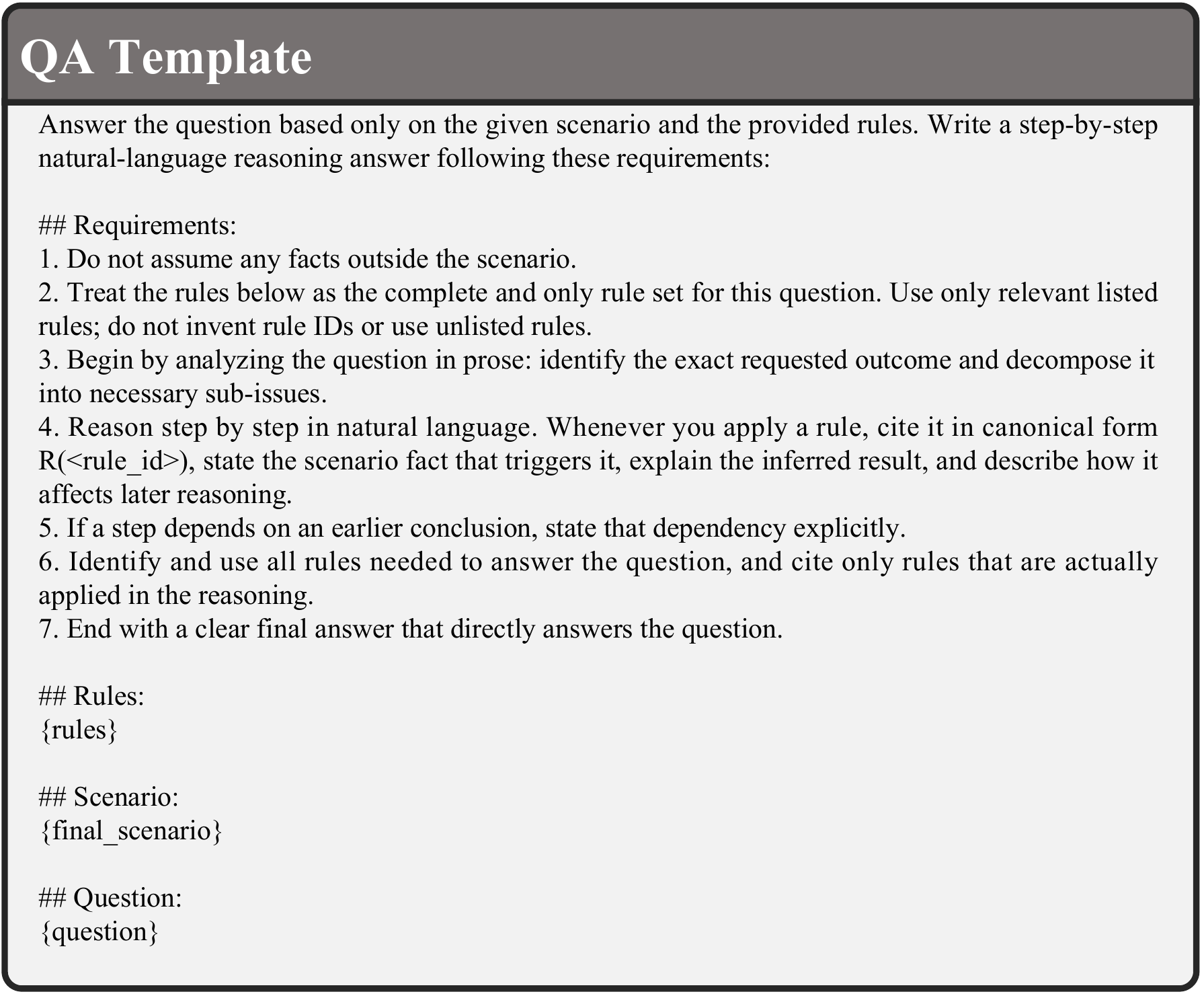}
    \caption{Question answering template.}
    \label{fig:question_answering_prompt}
\end{figure}

\begin{figure}[htbp]
    \centering
    \includegraphics[scale=0.23]{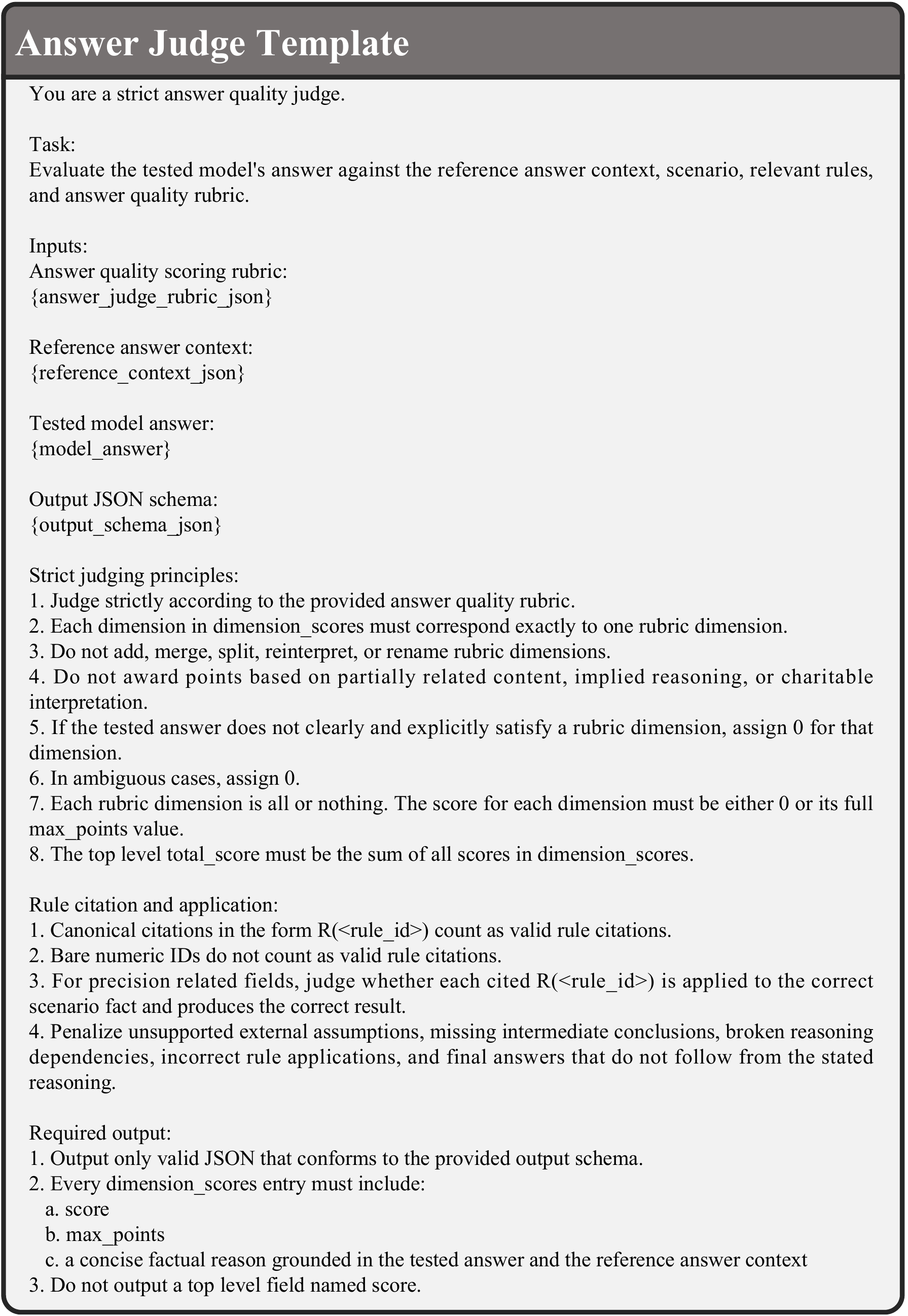}
    \caption{Answer judge template.}
    \label{fig:QA_template_answer_judge}
\end{figure}

\end{document}